\documentclass[12pt]{article}

\newcommand{\argmin}{\mathop{\arg\min}}
\newcommand{\pprime}{{\prime\prime}}
\newcommand{\dd}{\text{d}}

\newcommand{\vertiii}[1]{{\left\vert\kern-0.25ex\left\vert\kern-0.25ex\left\vert #1 
    \right\vert\kern-0.25ex\right\vert\kern-0.25ex\right\vert}}

\newcommand{\ReLU}{\mathrm{ReLU}}
\newcommand{\Lip}{\text{Lip}}
\newcommand{\id}{\text{id}} 
\newcommand{\Pdim}{\operatorname{Pdim}}

\newcommand{\Acal}{\mathcal{A}}

\newcommand{\Fcal}{\mathcal{F}}
\newcommand{\Gcal}{\mathcal{G}}
\newcommand{\Hcal}{\mathcal{H}}

\newcommand{\Lcal}{\mathcal{L}}

\newcommand{\Pcal}{\mathcal{P}}

\newcommand{\Vcal}{\mathcal{V}}

\newcommand{\Xcal}{\mathcal{X}}
\newcommand{\Ycal}{\mathcal{Y}}

\newcommand{\Pscr}{\mathscr{P}}

\newcommand{\EE}{\mathbb{E}}

\newcommand{\II}{\mathbb{I}}

\newcommand{\NN}{\mathbb{N}}

\newcommand{\PP}{\mathbb{P}}
\newcommand{\QQ}{\mathbb{Q}}
\newcommand{\RR}{\mathbb{R}}

\newcommand{\mfrak}{\mathfrak{m}}

\newcommand{\sfrak}{\mathfrak{s}}

\newcommand{\Abf}{\mathbf{A}}

\newcommand{\bbf}{\mathbf{b}}

\newcommand{\xbf}{\mathbf{x}}
\newcommand{\ybf}{\mathbf{y}}

\newcommand{\alphabf}{{\boldsymbol\alpha}}

\newcommand{\Drm}{\mathrm{D}}

\usepackage{amsthm,amsmath,amssymb,multirow,subfigure,makecell,booktabs,array,url,bm,mathtools,wrapfig,lipsum,mathrsfs,relsize,titling,dsfont,epstopdf,bbm,color,caption, comment,enumitem, geometry, tikz}
\usepackage{algorithm, algpseudocode}
\usepackage{graphicx}

\usepackage[export]{adjustbox}[2011/08/13]

\setlist{leftmargin=5mm}
\usepackage[colorlinks,linkcolor=red,citecolor=blue,urlcolor=blue]{hyperref}

\newtheoremstyle{mystyle}{}{}{}{}{\bfseries}{.}{ }
{\thmname{#1}\thmnumber{ #2}\thmnote{ (#3)}}
\theoremstyle{mystyle}
\newtheorem{theorem}{Theorem}[section] 
\newtheorem*{theorem*}{Theorem}
\newtheorem{definition}[theorem]{Definition}

\newtheorem{proposition}[theorem]{Proposition}
\newtheorem{assumption}[theorem]{Assumption}

\usepackage{thm-restate, cleveref}
\usepackage[page,toc,titletoc,title]{appendix}

\makeatletter
\def\keywords{\xdef\@thefnmark{}\@footnotetext}
\makeatother

\usepackage[backend=biber, style=alphabetic, sorting=nty]{biblatex}
\usepackage{titletoc}

\newcommand{\worktitle}{Mitigating the Curse of Dimensionality in Uniform Convergence of Deep Neural Networks via Smooth Activations}

\begin{document}

\title{\worktitle}
\author{\large Yizhe Ding, Runze Li, Jia Liu and Lingzhou Xue}
\date{\large Department of Statistics, The Pennsylvania State University}

\maketitle

\begin{abstract}

This paper establishes a theoretical framework for the uniform convergence of smoothly activated deep neural network (DNN) estimators. While standard ReLU networks achieve minimax-optimal rates in the $L^2(P)$ norm for various nonparametric regression tasks, we establish a theoretical lower bound demonstrating that least-squares ReLU estimators can suffer from the \textit{curse of dimensionality} in their uniform convergence behavior. Motivated by the need for reliable uniform guarantees in downstream tasks requiring worst-case reliability, we address this limitation by analyzing smoothly activated DNNs (smooth DNNs), encompassing both feedforward and residual structures. We establish novel pseudo-dimension bounds, non-asymptotic approximation guarantees, and Hölder-norm bounds for the approximators of these models. Leveraging these results, we derive non-asymptotic uniform convergence rates for smooth DNN estimators across multiple statistical contexts, including Huber, least-squares, quantile, and logistic regression. We prove that smooth DNNs can mitigate the {curse of dimensionality} in uniform convergence by adaptively exploiting the low-dimensional hierarchical composition structure of the target function. Supported by both simulation studies and a real-world application, our results position smooth DNNs as a theoretically grounded and practically viable alternative to ReLU networks for statistical learning tasks requiring uniform guarantees.

\end{abstract}


\section{Introduction}

Feedforward neural networks (FNNs) with rectified linear unit (ReLU) activations have received much attention in nonparametric regression due to their ability to efficiently approximate functions with latent low-dimensional structures~\cite{bauer2019deep, schmidt2020nonparametric, kohler2021rate}. This structural flexibility has facilitated their applications to a broad range of statistical problems, including 
robust regression~\cite{fan2024noise,ding2025new}, 
survival analysis~\cite{zhong2022deep}, and factor and interaction models~\cite{fan2024factor, bhattacharya2024deep},  among others. 

While ReLU FNN estimators achieve minimax-optimal convergence rates in the $L^2(P)$ norm~\cite{fan2024noise}, their uniform convergence properties remain less understood. Yet, uniform convergence is a fundamental prerequisite for many downstream statistical theoretical analyses and decision-making applications, such as ensuring reliable individualized treatment recommendations~\cite{guo2021estimation}, constructing valid confidence bands~\cite{yao2005functional}, and enabling transfer learning~\cite{schmidt2024local}. Current theoretical guarantees for the uniform convergence of ReLU FNNs are notably limited: they are often restricted to shallow ReLU network estimators with univariate covariates \cite{schmidt2024local} or rely on the availability of a uniformly consistent pilot estimator~\cite{imaizumi2023sup}. This critical theoretical gap raises significant concerns regarding the reliability of ReLU FNN estimators in contexts where uniform guarantees are essential, which restricts their practical applicability. 

To investigate this open problem, we first characterize a fundamental bottleneck in the uniform convergence of ReLU FNNs. Specifically, we establish the first theoretical lower bound in the literature demonstrating that least-squares ReLU FNN estimators inherently suffer from the \textit{curse of dimensionality} in their uniform convergence rates, even when they simultaneously achieve the minimax-optimal convergence rate in the $L^2(P)$ norm. This underscores that the optimal $L^2(P)$ convergence of ReLU FNNs does not translate to the uniform reliability required for downstream statistical theoretical analyses.

Beyond these theoretical limitations, the focus on ReLU FNNs increasingly diverges from modern deep learning practice. The piecewise linear ReLU activation has largely been replaced by $C^\infty$–smooth alternatives, such as the sigmoid linear unit (SiLU)~\cite{ramachandran2017swish, elfwing2018silu}, and the Gaussian error linear unit (GELU)~\cite{hendrycks2016gelu}, driven by their superior empirical performance. Furthermore, residual architectures (ResNets)~\cite{he2015resnet, he2016identity} have fundamentally reshaped neural network design. By resolving the vanishing gradient and degradation problems inherent in training very deep networks, residual connections have become the indispensable backbone of modern deep learning. Although the statistical theory underlying these components remains limited, the integration of residual architectures with $C^\infty$-smooth activations forms the structural basis of current powerful foundation models, from Vision Transformers~\cite{dosovitskiy2021vit} to state-of-the-art large language models like LLaMA~\cite{touvron2023llama} and DeepSeek~\cite{xie2025mhc}. 

To overcome this bottleneck, motivated by recent architectural developments in deep learning, we focus on deep neural networks with smooth activations (smooth DNNs), encompassing both feedforward and residual architectures, and develop a rigorous theoretical framework for their uniform convergence analysis. Through a comprehensive theoretical analysis together with supporting simulation studies, we show that smooth DNNs can substantially mitigate the \textit{curse of dimensionality} in uniform convergence and provide strictly stronger uniform guarantees than ReLU FNNs. These results position smooth DNNs as a principled and theoretically grounded alternative for statistical learning tasks in which uniform convergence is essential.

The primary methodological and theoretical contributions of this work are threefold. Together, they bridge a critical gap between the empirical success of smooth DNNs and the current lack of a general statistical theory for their uniform convergence:
\begin{itemize}
    \item \textbf{Characterizing the {curse of dimensionality} for ReLU FNNs in the uniform convergence}. We establish the first theoretical lower bound demonstrating that ReLU FNNs inherently suffer from the \textit{curse of dimensionality} in the uniform norm. While achieving minimax-optimal $L^2(P)$ convergence~\cite{fan2024noise}, Theorem~\ref{thm: lower bound of ReLU with penalty} reveals that the uniform convergence rate of the least-squares ReLU FNN estimator can be bounded below by $n^{-\frac{1}{d+1}}$ and above by $n^{-\frac{1}{d+2}}$, where $n$ is the sample size and $d$ is the feature dimension, regardless of the Hölder smoothness of the target function. Furthermore, our analysis based on interpolation inequalities in Section~\ref{sec: Origin of Uniform Convergence}  suggests that this limitation stems fundamentally from the limited smoothness of the ReLU activation,  which prevents ReLU FNNs from exploiting higher-order regularity.

    \item \textbf{A foundational theoretical framework for smooth DNNs.} The statistical theory for deep neural networks, particularly ResNets, with smooth activations, remains largely absent. Existing literature predominantly focuses on either FNNs with less common activations, such as sigmoid, tanh, or rectified power unit (RePU) activations~\cite{de2021approximation, belomestny2023simultaneous, shen2023differentiable} or residual architectures with ReLU activations~\cite{oono2019approximation, liu2021besov, liu2022benefits}. To enable the uniform convergence analysis of smooth DNNs, Section~\ref{sec: Statistical Properties of DSNNs} develops a comprehensive set of theoretical tools. Specifically, we establish an upper bound on the pseudo-dimension in Theorem~\ref{thm: pseudo-dimension of Pfaffian network} and derive approximation error bounds for both Sobolev functions and hierarchical composition models in Theorems~\ref{thm: approximation error and sobolev bound on Omega for sobolev smooth function} and~\ref{thm: approximation error and holder bound on Omega for hierarchical composition model}, respectively. To our knowledge, these results provide the first theoretical foundation for establishing uniform convergence guarantees for smooth DNN estimators.

    \item \textbf{Uniform convergence and robustness guarantees.} Building on our theoretical framework, we establish uniform convergence guarantees for smooth DNN estimators in Huber, least-squares, quantile, and logistic regression. In particular, Theorem~\ref{thm: uniform convergence rate of deep smooth Huber regression} shows that the smooth DNN Huber estimator is non-asymptotically robust in the uniform norm, extending existing robustness results for ReLU FNN Huber estimators in the $L^2(P)$ norm~\cite{fan2024noise, ding2025new}. Theorem~\ref{theorem: uniform convergence rate of deep smooth LS regression} then gives the corresponding uniform convergence rate for the smooth DNN least-squares estimator as a special case of Huber regression, and shows that it can overcome the \textit{curse of dimensionality} exhibited by ReLU least-squares regression in Theorem~\ref{thm: lower bound of ReLU with penalty}. Analogous uniform convergence and robustness guarantees for quantile and logistic regression are established in Theorems~\ref{thm: uniform convergence rate for DSNN quantile regression} and~\ref{theorem: uniform convergence rate for DSNN logistic regression}, respectively. Notably, the uniform convergence guarantee for logistic regression also provides rigorous theoretical support for downstream applications such as probability estimation for double Higgs boson production~\cite{manole2024background}; see Theorem~\ref{theorem: uniform convergence rate for density-ratio based distribution estimate}. More broadly, across all these tasks, we prove that smooth DNN estimators can adapt to the low-dimensional hierarchical composition structure of the target function, thereby enjoying a clear theoretical advantage over ReLU FNNs in uniform convergence.
\end{itemize}

To support our methodological and theoretical results, Section~\ref{sec: Numerical Studies} benchmarks the numerical $L^2(P)$ and uniform estimation errors of smooth DNN estimators against those of ReLU FNN estimators in Huber regression. The substantial performance gains achieved by smooth DNN estimators provide empirical support for our methods and theory. Taken together, these results support smooth DNN estimators as effective alternatives to ReLU FNNs in practical applications where uniform convergence is required.

The rest of this paper is organized as follows. 
Section~\ref{sec: Limitations of ReLU Networks in Uniform Convergence} establishes the lower and upper uniform convergence rates of the least squares ReLU FNN estimators. Section~\ref{sec: Methodology: deep smooth residual networks} presents the new methodology for smooth DNNs. 
In Section~\ref{sec: Statistical Properties of DSNNs}, we establish the key statistical properties of smooth DNNs. 
Building on these foundational elements, in Section~\ref{sec: Applications of DSNNs}, we derive uniform convergence rates of smooth DNN estimators for Huber, least-squares, quantile, and logistic regression. 
In Section~\ref{sec: Numerical Studies}, we present simulation studies and a real application. 
Section~\ref{sec: Conclusion} includes a few concluding remarks. The complete proofs and additional technical or numerical results are presented in the Supplementary Materials.

Before proceeding, we introduce the following notation and terminology that will be used throughout this paper. We use the notation $a\lesssim b$ to mean there is a constant $C>0$ independent of $a$ and $b$, such that $a\leq C b$. We say $a\asymp b$ if both $a\lesssim b$ and $b\lesssim a$ hold. We say $a\lesssim_{\log n}b$ if there exists a function $C(\log n)$ such that $a\leq C(\log n)\cdot b$. We may further use the notation $\lesssim_{\log}$ to suppress any other logarithmic terms. $a\lor b = \max(a,b)$ and $a\land b = \min(a,b)$. For $n\in\NN_+$, denote $[n]=\{1,2,\cdots,n\}$. For $x\in \RR$, denote $\lceil x\rceil=\min\{n\in\NN: n\geq x\}$. For $x\in\RR^d$, denote its $\ell_2$ norm as $\|\cdot\|_2$ and $\ell_\infty$ norm as $\|\cdot\|_\infty$. For any $M>0$, let $L^\infty(M)$ denote the collection of real-valued functions uniformly bounded by $M$. For $\Omega\subset \RR^d$, we denote the uniform norm of a function $f$ on $\Omega$ by $\|f\|_{L^\infty(\Omega)}$. For $\xbf=(x_1,\cdots,x_d)\in\RR^d$ and $\alphabf=(\alpha_1,\cdots,\alpha_d)\in\NN^d$, we denote $\xbf^\alphabf=x_1^{\alpha_1}x_2^{\alpha_2}\cdots x_d^{\alpha_d}$, and $|\alphabf|=|\alpha_1|+|\alpha_2|+\cdots+|\alpha_d|$. Let $\Drm_i$ be the derivative operator to the $i$-th variable for $i\in[d]$, the multi-index derivative of order $\alphabf\in\NN^d$ is defined as $\Drm^\alphabf:=\Drm_{1}^{\alpha_1}\cdots \Drm_{d}^{\alpha_d}$. Let $\beta=r+s$ for some non-negative integer $r$ and $0<s\leq 1$, $d\in\NN_+$, and $C>0$. Let $\|\cdot\|_{C^\beta(\Omega)}$ denote the Hölder norm of order $\beta$, and a $d$-variate function $f$ is called $(\beta,C)$-smooth on $\Omega\subseteq\RR^d$, if 
\[
    \|f\|_{C^\beta(\Omega)}:=\max_{0\leq |\alphabf|\leq r}\|\Drm^\alphabf f\|_{L^\infty(\Omega)}\lor\max_{\alphabf:|\alphabf|=r}\sup_{\xbf,\ybf\in\Omega, \xbf\neq \ybf}\frac{|\Drm^\alphabf f(\xbf)-\Drm^\alphabf f(\ybf)|}{\|\xbf-\ybf\|_2^s}\leq C.
\]

\section{Uniform Convergence of ReLU FNN Estimators}\label{sec: Limitations of ReLU Networks in Uniform Convergence}

In this section, we investigate the fundamental limitations of ReLU FNN estimators in achieving uniform convergence. Section~\ref{sec: Lower Bound for Uniform Convergence of ReLU Estimators} establishes a theoretical lower bound on the uniform convergence rates of least-squares ReLU estimators in Theorem~\ref{thm: lower bound of ReLU with penalty}, demonstrating that they inherently suffer from the \emph{curse of dimensionality}, even when their $L^2(P)$ convergence rate is minimax-optimal. To explain the mechanism behind this bottleneck, Section~\ref{sec: Origin of Uniform Convergence} reveals how the limited smoothness of ReLU FNNs leads to slow uniform convergence rates.

\subsection{A Lower Bound on Uniform Convergence Rates}\label{sec: Lower Bound for Uniform Convergence of ReLU Estimators}

To establish that the \emph{curse of dimensionality} in uniform convergence is an inherent limitation of ReLU FNN estimators, we focus on the foundational framework of nonparametric least-squares regression. Let $\{(X_i,Y_i)\}_{i=1}^n$ be the \emph{i.i.d.}\ observations generated from
\begin{equation}\label{eq: definition of least squares regression}
    Y_i = f_0(X_i) + \xi_i, 
\end{equation}
where $f_0$ is the unknown regression function, $\xi_i$ denotes the random noise, and $P$ represents the distribution of $(X_i,Y_i)$. 

To estimate $f_0$, we consider empirical risk minimization over a class of ReLU FNNs. Specifically, given a network width $W\in\NN_+$ and depth $D\in\NN_+$, a ReLU FNN is defined as a function $f:\RR^d\to\RR$ of the composition form
\begin{equation}\label{eq: definition of feedforward network}
    f(x) = \Lcal_{D+1}\circ\sigma_\ast\circ\Lcal_{D}\circ\sigma_\ast\circ\cdots\circ\Lcal_{2}\circ\sigma_\ast\circ\Lcal_{1}(x),
\end{equation}
where $\sigma_\ast(x)=x\lor 0$ is the ReLU activation function applied entrywise, and $\Lcal_i:\RR^{d_{i-1}}\to \RR^{d_i}$ are affine transformation maps with the dimension vector $(d_0,\ldots,d_{D+1})=(d,W,\ldots,W,1)$.

To rule out pathological estimators that interpolate the noise through increasingly steep local oscillations, we restrict attention to uniformly bounded ReLU networks with uniformly bounded parameters. Accordingly, for some $r\ge 2$ and $M\ge 0$, we define the candidate class by
\begin{equation*}
\begin{aligned}
    \Fcal_{\ReLU}&:=\bigl\{f\text{ of the form~\eqref{eq: definition of feedforward network}: }\max_{i\in[D+1]}\|A_i\|_{\infty}\lor \|b_i\|_{\infty}\leq r\bigr\}\cap L^\infty(M).
\end{aligned}
\end{equation*}
Let the corresponding least-squares ReLU estimator $\hat f_n$ be defined by
\begin{equation}\label{eq: definition of ReLU least squares estimator}
    \hat f_n \in \argmin_{f\in\Fcal_{\ReLU}} \sum_{i=1}^n \bigl(f(X_i)-Y_i\bigr)^2,
\end{equation}
We further require that the estimator sequence is uniformly Lipschitz, in the sense that $\sup_{n\ge 1}\|\hat f_n\|_{\Lip}<\infty$. 
These restrictions are imposed to exclude the possibility that poor uniform convergence arises from pathological overfitting to the noise.

Meanwhile, to isolate the architectural limitation of ReLU FNNs from pathological distributions, we also impose standard regularity conditions on the data-generating process. Specifically, it is assumed in Theorem~\ref{thm: lower bound of ReLU with penalty} below that the covariates $X_i$ are uniformly distributed on $[0,1]^d$, the noises $\xi_i$ are independent of $X_i$ and uniformly bounded, and the target function $f_0$ is H\"older-smooth and uniformly bounded. These assumptions describe a highly regular nonparametric estimation setting: the covariate distribution is benign, the regression function is smooth, and the noise is bounded. Therefore, any slow uniform convergence rate established below cannot be attributed to irregular features of the data-generating process.

Taken together, the restrictions on the estimator and on the data-generating distribution ensure that the lower bound in Theorem~\ref{thm: lower bound of ReLU with penalty} reflects an intrinsic limitation of the ReLU architecture itself. In particular, the slow uniform convergence rate is not caused by heavy-tailed noise, pathological covariate design, lack of smoothness of $f_0$, or overfitting to the observations, but rather by the intrinsic limitations of ReLU FNN architectures. Under these conditions, we obtain the following lower bound on the uniform convergence rate of $\hat f_n$.

\begin{theorem}\label{thm: lower bound of ReLU with penalty}
    Let $\Pcal$ denote the collection of distributions $(X_i,Y_i)\sim P$ such that, in model~\eqref{eq: definition of least squares regression}, the covariates $X_i$ follow uniform distribution on $[0,1]^d$, the noise satisfies $|\xi_i|\leq 1$ and $\EE[\xi_i]=0$, and the regression function satisfies $f_0\in C^\alpha([0,1]^d)\cap L^\infty([0,1]^d)$ for some $\alpha\geq 2$. Suppose the ReLU FNN with width $W$ and depth $D$ satisfies, $n^{\frac{3}{4(d+1)}} \lesssim W \lesssim D \lesssim 2^{\,n^{\frac{3}{16(d+1)}}}$ up to poly-logarithmic factors. Then for any $d\geq 2$, 
    \[
        \sup_{P\in\Pcal}\ 
        \EE\bigl\|\hat f_n-f_0\bigr\|_{L^\infty([0,1]^d)}\ \gtrsim\ n^{-\frac{1}{d+1}}.
    \]
\end{theorem}

Theorem~\ref{thm: lower bound of ReLU with penalty} explicitly formalizes the classical \emph{curse of dimensionality} for the least-squares ReLU estimator  $\hat{f}_n$, since the exponent of its uniform convergence rate depends poorly on dimension $d$. In contrast, \cite{fan2024noise} proved that the $L^2(P)$ convergence rates of ReLU FNN estimators do not have this issue. Specifically, under these conditions, the $L^2(P)$ estimation error of ReLU estimator $\hat{f}_n$ satisfies
\[
    \sup_{P\in\Pcal}\EE\|\hat f_n-f_0\|_{L^2(P)}
    \ \lesssim_{\log}\ 
    \sqrt{\frac{(WD)^2}{n}} \;+\; (WD)^{-2\alpha/d}.
\]
In particular, provided the target function $f_0$ is not excessively smooth (i.e., $\alpha \le d(d-2)/6$), one can choose a network architecture with $W\asymp D \asymp n^{\frac{d}{4(d+2\alpha)}} \geq n^{\frac{3}{4(d+1)}}$. Under this configuration, $\hat f_n$ simultaneously attains the minimax-optimal $L^2(P)$ rate $n^{-\frac{\alpha}{2\alpha+d}}$ while remaining trapped at the slow, dimension-dependent rate of $n^{-\frac{1}{d+1}}$ in the uniform norm. This stark discrepancy highlights a critical intrinsic limitation that least-squares ReLU FNN estimators can be suboptimal for downstream applications requiring uniform convergence guarantees. We provide a more detailed discussion of Theorem~\ref{thm: lower bound of ReLU with penalty} and its implications in Section~S1 of the Supplementary Material.

\subsection{An Interpolation Perspective on Uniform Convergence Rates}\label{sec: Origin of Uniform Convergence}

In Section~\ref{sec: Lower Bound for Uniform Convergence of ReLU Estimators}, ReLU least squares estimators are shown to suffer from the \textit{curse of dimensionality} under the uniform norm. In contrast, many classical nonparametric estimators, such as splines, wavelets, and Fourier series, typically avoid this issue. A key reason is their linear structure: these estimators lie in the linear span of basis functions and, conditioned on the design matrix, behave analogously to ordinary least squares (OLS) estimators.

To illustrate this structural contrast, consider observations $\{(X_i,Y_i)\}_{i=1}^n$ generated from the parametric linear model $Y_i = X_i^\top \beta_0 + \xi_i$, where $X_i$ is supported on $[0,1]^d$, and let $\hat\beta_n$ denote the OLS estimator. Define the true and estimated regression functions by $f_0(x) = x^\top \beta_0$ and $\hat f_n(x) = x^\top \hat\beta_n$, respectively. Assuming the population second-moment matrix $\Sigma:=\EE[XX^\top]$ is positive definite with the smallest eigenvalue $\lambda_{\min}(\Sigma)>0$, the $L^2(P)$ and $L^\infty$ estimation errors are deterministically linked. Specifically, because $\|x\|_2 \le \sqrt{d}$, Cauchy--Schwarz implies $\|\hat f_n - f_0\|_{L^\infty([0,1]^d)} \le \sqrt{d} \|\hat\beta_n - \beta_0\|_2$. Concurrently, $\|\hat f_n - f_0\|_{L^2(P)}^2 = (\hat\beta_n - \beta_0)^\top \Sigma (\hat\beta_n - \beta_0) \ge \lambda_{\min}(\Sigma) \|\hat\beta_n - \beta_0\|_2^2$. Combining these relations yields the inequality that implies that the $L^\infty$ and $L^2(P)$ estimation errors are equivalent for linear models: 
\[
    \sqrt{\frac{\lambda_{\min}(\Sigma)}{d}} \|\hat f_n - f_0\|_{L^\infty([0,1]^d)}
    \leq \|\hat f_n - f_0\|_{L^2(P)}
    \leq \|\hat f_n - f_0\|_{L^\infty([0,1]^d)}.
\]

Beyond parametric linear models, analogous interpolation inequalities serve as powerful classical tools to establish uniform convergence rates for broader function classes by exploiting underlying smoothness properties. For example, the Gagliardo--Nirenberg inequality~\cite{gagliardo1959ulteriori, nirenberg1966extended} applies to Hölder functions, while the Brezis--Mironescu inequality~\cite{brezis2018gagliardo} applies to (fractional) Sobolev functions. 

This viewpoint provides an analytic explanation for the slow uniform convergence rates observed in ReLU FNN estimators. To see this, consider the standard Gagliardo--Nirenberg inequality on $[0,1]^d$: for any function $f \in C^\alpha([0,1]^d)$, the uniform norm is bounded by 
\begin{equation}\label{eq: Interpolation inequality for Hölder-smooth functions}
    \|f\|_{L^\infty([0,1]^d)} \lesssim \|f\|_{C^\alpha([0,1]^d)}^{\frac{d}{2\alpha+d}} \cdot \|f\|_{L^2(P)}^{\frac{2\alpha}{2\alpha+d}}.
\end{equation}
This interpolation bound shows that when the candidate function class possesses substantial smoothness ($\alpha \gg d$), the $L^\infty$ convergence rate closely tracks the $L^2(P)$ rate. However, because ReLU FNNs are generally only piecewise affine and need not belong to $C^1$, this interpolation argument can use at most Hölder exponents $\alpha\in(0,1)$. Applying \eqref{eq: Interpolation inequality for Hölder-smooth functions} together with the standard $L^2(P)$ rate $\|\hat f_n - f_0\|_{L^2(P)} \lesssim n^{-1/2}$, and then letting $\alpha\to1$, yields
\[
    \|\hat f_n - f_0\|_{L^\infty([0,1]^d)} \lesssim n^{-1/(d+2)}.
\]
When the Gagliardo--Nirenberg inequality is tight, this bound is sharp. This issue shows that the poor intrinsic smoothness of the ReLU activation function fundamentally prevents its network architecture from adapting to fast uniform convergence requirements, rendering it ill-suited for downstream tasks requiring strong uniform convergence guarantees.

\section{Smoothly Activated Deep Neural Networks}\label{sec: Methodology: deep smooth residual networks}

This section introduces smoothly activated deep neural networks (smooth DNNs), which encompass both deep feedforward and deep residual architectures, as a powerful and theoretically rigorous alternative. Section~\ref{sec: Smooth Activation Functions} reviews smooth activation functions widely adopted in modern deep learning and then summarizes the foundational framework of residual networks. Leveraging these components, Section~\ref{sec: Deep Smooth Residual Networks} presents the mathematical formulation and characterization of smooth DNNs.

\subsection{Preliminaries}\label{sec: Smooth Activation Functions} We first characterize the infinitely differentiable (i.e., $C^\infty$-smooth) activation functions and the structural configurations of residual networks.

\textbf{Smooth Activation Functions.} Nonlinear activation functions govern the expressiveness of neural networks. While the ReLU activation serves as a baseline for deep compositional mapping, its non-differentiability at the origin constrains its functional regularity. Non-smooth variants such as Leaky ReLU, PReLU, ELU, and SELU introduce non-zero components on the negative real line but remain non-differentiable, making them analytically ill-suited for establishing fast uniform convergence rates under the interpolation perspective described in Section~\ref{sec: Origin of Uniform Convergence}. 

\begin{figure}[ht]
    \centering
    \includegraphics[width=0.8\linewidth]{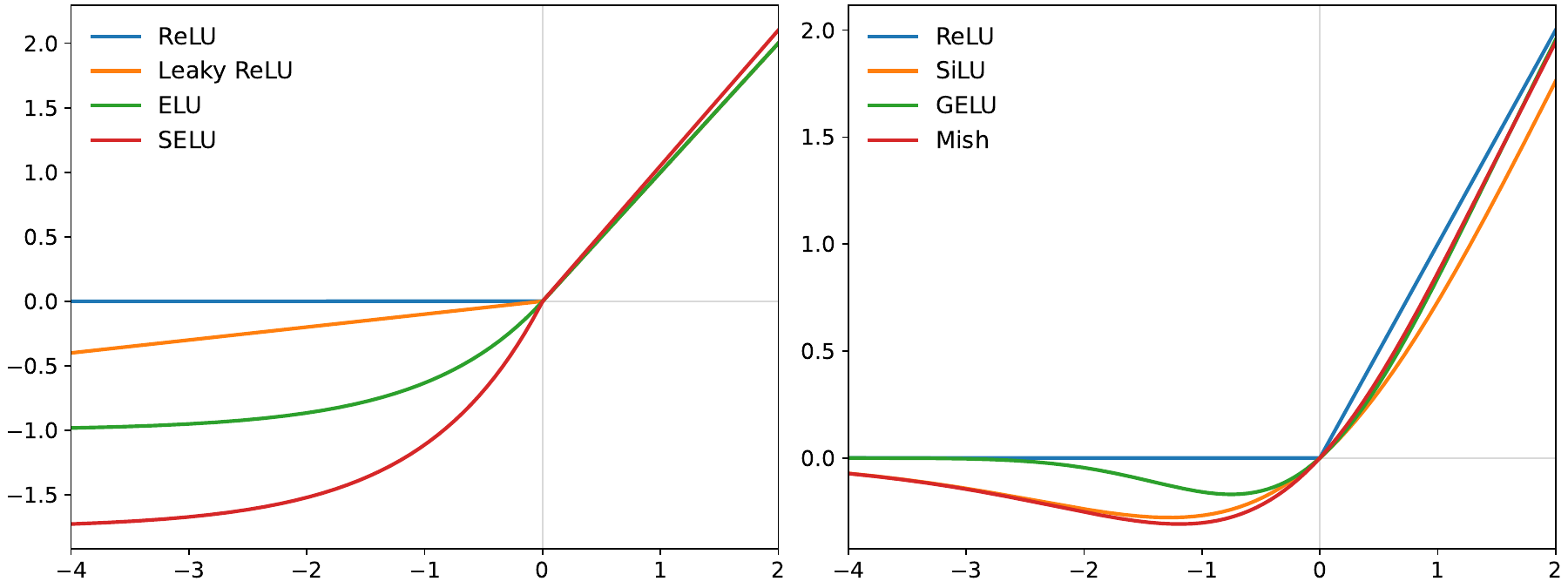}
    \caption{ReLU and its non-$C^\infty$ variants (left); and its $C^\infty$ variants (right).}
    \label{fig: all activations}
\end{figure}

To achieve higher-order functional smoothness, we consider a class of $C^\infty$-smooth activations, including SiLU (also known as Swish)~\cite{ramachandran2017swish, elfwing2018silu}, GELU~\cite{hendrycks2016gelu}, and Mish~\cite{misra2019mish}. As illustrated in Figure~\ref{fig: all activations}, these activations exhibit smooth transitions near the origin. 
Formally, these $C^\infty$-smooth activation functions admits a factorization of the form $\sigma(x)=x\cdot \psi(x)$, where $\psi: \RR \to \RR$ is a non-decreasing, uniformly bounded, and infinitely differentiable function. The explicit functional forms are given by: 
\begin{align}
    \text{SiLU}(x) =& x\cdot \frac{\exp(x)}{1+\exp(x)}; \label{eq:SiLU}\\
    \text{GELU}(x) =& x\cdot \frac{1}{2}\Bigl(1+\tanh\bigl(\sqrt{2/\pi}\,(x+0.044715\,x^3)\bigr)\Bigr); \label{eq:GELU}\\
    \text{Mish}(x) =& x\cdot \tanh\!\bigl(\log(1+\exp(x))\bigr). \label{eq:Mish}
\end{align}
Here we use the hyperbolic tangent version of the GELU activation, and its associated function $\psi$ approximates the distribution function of the standard normal distribution. Writing $\ReLU(x)=x\cdot H(x)$ with $H(x)=\II(x\geq 0)$, the corresponding functions $\psi$ of SiLU, GELU, and Mish converge to $H(x)$ exponentially fast as $x\to\pm\infty$. As shown in Section~\ref{sec: Approximation Error}, this property ensures that the resulting network classes enjoy strong approximation properties for functions in Sobolev spaces and hierarchical composition models.

\textbf{Residual Networks.} 
In the context of nonparametric function estimation, ReLU FNNs have been shown to possess strong approximation capabilities relative to classical nonparametric models, such as wavelets~\cite{bauer2019deep, schmidt2020nonparametric, kohler2021rate}. However, as compositional depth increases, standard deep feedforward architectures often exhibit ill-conditioned optimization landscapes, which can lead to higher empirical risk even under careful parameter initialization and normalization strategies~\cite{he2015resnet}.

To regularize the optimization landscape, \emph{residual networks} (ResNets) incorporate identity skip connections around each affine--activation block~\cite{he2015resnet,he2016identity}. Rather than applying an isolated nonlinear transformation at each layer, the residual architecture models a functional perturbation of the preceding state. Formally, given the output of the $\ell$-th layer $h_{\ell}(x)$, the $(\ell+1)$-th layer is defined as $h_{\ell+1}(x)\;=\;h_{\ell}(x)\;+\;\Fcal_\ell\bigl(h_{\ell}(x)\bigr)$, where $\Fcal_\ell$ is typically implemented via a small stack of affine layers, normalizations, and nonlinear activations.

\begin{figure}[ht]
    \centering
    \includegraphics[width=\linewidth]{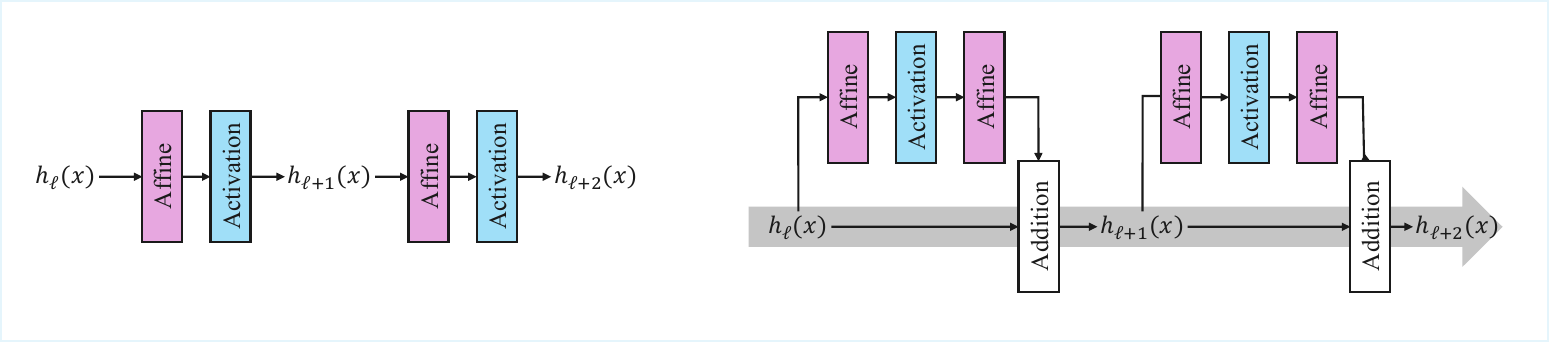}
    \caption{Architectures of an FNN (left) and residual blocks with $\Lcal_1=\id$ (right).}
    \label{fig:FNN_and_ResNet}
\end{figure}

As illustrated in Figure~\ref{fig:FNN_and_ResNet}, the identity skip connection creates a direct pathway $h_{\ell}\!\to h_{\ell+1}$ for more stable information and gradient propagation across layers. By mitigating the structural degradation of gradients, residual architectures make DNNs easier to train. We will incorporate residual architectures into our theoretical framework as an optimization-oriented generalization of feedforward architectures.

\subsection{Mathematical Formulation of Smooth DNNs}\label{sec: Deep Smooth Residual Networks}

We now define the function classes of smooth DNNs under both feedforward and residual architectures. To parameterize these architectures simultaneously, we first introduce an activation operator that unifies coordinatewise mappings and Gated Linear Unit (GLU) formulations.

\begin{definition}[Activation operator]\label{def: activation operator}
    Given a $C^\infty$-smooth activation function $\sigma:\RR\to\RR$, the activation operator $\Acal_\sigma:\RR^q\to\RR^{\tilde q}$ is defined by one of the following mappings:
    \begin{itemize}
        \item \textbf{Standard activation:} $q=\tilde q$, and $\Acal_\sigma(z)=\sigma(z)$, where $\sigma(z)$ is evaluated coordinatewise.
        \item \textbf{GLU-type activation:} $q=2\tilde q$, and for $z=(z_1^\top,z_2^\top)^\top\in\RR^q$ with $z_1,z_2\in\RR^{\tilde q}$,
        \(
            \Acal_\sigma(z)=\sigma(z_1)\odot z_2,
        \)
        where $\sigma(z_1)$ is evaluated coordinatewise and $\odot$ denotes the Hadamard product.
    \end{itemize}
\end{definition}

\begin{definition}[Residual block]\label{def: Residual block}
    Given the input and output dimensions $d_1,d_2\in\NN_+$ and width $W_1,W_2\in\NN_+$, let $q = W_2$ for standard activation, and $q=2W_2$ for GLU-type activation. Denote matrices $\Abf_1\in \RR^{W_1\times d_1}$, $\Abf_2\in\RR^{q\times W_1}$, and $\Abf_3\in \RR^{d_2\times  W_2}$, vectors $\bbf_1\in\RR^{W_1}$, $\bbf_2\in\RR^{q}$, and $\bbf_3\in\RR^{d_2}$, and a $C^\infty$-smooth activation function $\sigma$, a residual block $\phi$ with parameters $(d_1,d_2,W_1,W_2)$ is defined as
    \[
        \phi(x) = (\id + \Lcal_3\circ \Acal_\sigma\circ\Lcal_2)\circ \Lcal_1(x),\qquad x\in\RR^{d_1},
    \]
    where $\id$ denotes the identity mapping and $\Lcal_i(z):= \Abf_i z + \bbf_i$ for $i=1,2,3$. 
\end{definition}

Next, we define smooth DNNs for both residual and feedforward architectures.

\begin{definition}[Smooth DNN]\label{def: resnet}
    A \emph{smooth ResNet} of depth $D$ and width $W$ is a function $f:\RR^{d_1}\to\RR^{d_{D+2}}$ of the form
    \begin{equation}\label{eq: definition of smooth rn}
        f(x)=\Lcal_{D+1}\circ \phi_D\circ\phi_{D-1}\circ\cdots\circ\phi_1(x),
    \end{equation}
    where each $\phi_i:\RR^{d_i}\to\RR^{d_{i+1}}$ is a residual block with parameters $(d_i, d_{i+1}, W_{1i}, W_{2i})$ as in Definition~\ref{def: Residual block}, and $\Lcal_{D+1}:\RR^{d_{D+1}}\to\RR^{d_{D+2}}$ is affine. The hidden dimensions of residual blocks are required to satisfy
    \(
        \Bigl(\max_{2\le i\le D+1} d_i\Bigr)\lor \Bigl(\max_{1\le i\le D}(W_{1i}\lor W_{2i})\Bigr)\le W.
    \)

    A \emph{smooth FNN} of depth $D$ and width $W$ is a function $f:\RR^{d_1}\to\RR^{d_{D+2}}$ of the form
    \begin{equation}\label{eq: definition of smooth fnn}
        f(x)=\Lcal_{D+1}\circ \Acal_\sigma\circ \Lcal_D\circ\cdots\circ \Acal_\sigma\circ \Lcal_1(x),
    \end{equation}
    where $\Lcal_i:\RR^{d_i}\to\RR^{\tilde{d}_{i}}$ are affine maps, with input dimensions given by $(d_1,d_2,\ldots,d_{D+1})=(d_1,W,\ldots,W)$. The output dimensions are given by $(\tilde{d}_1,\ldots,\tilde{d}_{D},\tilde{d}_{D+1})=(W,\ldots,W,d_{D+2})$ in the standard activation case, and $(2W,\ldots,2W,d_{D+2})$ in the GLU-type case.
\end{definition}

For a fixed activation function $\sigma \in C^\infty(\RR)$, the functional class of smooth DNNs with depth $D$, width $W$, and mapping from $\RR^{d_1}$ to $\RR^{d_{D+2}}$ is given by:
\[
    \Fcal(d_1,d_{D+2},D,W,\sigma) = \bigl\{x\mapsto f(x): f \text{ is of the form \eqref{eq: definition of smooth rn} or \eqref{eq: definition of smooth fnn}}\bigr\}.
\]
Unlike ReLU networks, the compositional structure of smooth DNNs ensures that any $f \in \Fcal(d_1, d_{D+2}, D, W, \sigma)$ inherits the $C^\infty$ regularity of its activation primitives, rendering the empirical process amenable to high-order metric entropy bounds. Apart from the architectural change, the estimation procedure remains the same as for ReLU networks. Specifically, given a loss function $\ell: \RR^{d_y} \times \RR^{d_y} \to \RR_+$ and \emph{i.i.d.}\ observations $\{X_i,Y_i\}\in \RR^{d_x}\times \RR^{d_y}$ for $i\in[n]$, we define the smooth DNN estimator by
\begin{equation}
    \hat{f}_n \;\in\; \argmin_{f\in\Fcal(d_x,d_y,D,W,\sigma)} \frac{1}{n}\sum_{i=1}^n \ell\big(f(X_i),\, Y_i\big).
\end{equation}

\section{Theoretical Properties of Smooth DNNs}\label{sec: Statistical Properties of DSNNs}

This section develops the core analytic properties of the smooth DNNs. Specifically, the pseudo–dimension (Pdim) is quantified in Section~\ref{sec: Pseudo-Dimension}. The approximation error bounds for Sobolev functions and hierarchical composition models are established in Section~\ref{sec: Approximation Error}. Hölder norm upper bounds for approximators are provided in Section~\ref{sec: Upper Bounds of Hölder Norms}.

\subsection{Pseudo-Dimension}\label{sec: Pseudo-Dimension}

In statistical learning theory, the statistical (generalization) error of $\hat{f}_n$ is governed by the complexity of the candidate class $\Fcal(d_1,d_{D+2},D,W,\sigma)$, with higher complexity typically yielding slower rates. Complexity analysis for neural networks began with binary threshold activations and was later extended to real-valued networks via the Pdim \cite{anthony2009neural}. More recently, nearly tight Pdim bounds have been established for networks with piecewise-polynomial activations, most notably ReLU \cite{bartlett2019nearly}. While these results largely tie complexity control to specific activation classes, \cite{goldberg1993bounding} bounded complexity in terms of the operation number needed to compute the network output. Building on this idea, \cite{karpinski1997polynomial} showed that when the neural network is describable by Pfaffian functions, the Pdim admits polynomial upper bounds in the number of parameters. This framework is particularly convenient for establishing Pdim bounds of smooth DNNs. Hence, we first recall the definition of a Pfaffian function.

\begin{definition}[Pfaffian Function]\label{def: Pfaffian Functions}
    A \textit{Pfaffian chain} of order $\ell\geq 0$ and degree $\alpha\geq 1$ on a domain $U\subseteq \RR^d$ is a sequence of analytic functions $(f_1,f_2,\cdots,f_\ell)$ over $U$ such that, 
    \[
        \frac{\dd f_j(\xbf)}{\dd x_i} = g_{ij}(\xbf,f_1(\xbf),\cdots,f_j(\xbf)),
        \quad i\in[d],\, j\in[\ell],\quad\text{for any $\xbf=(x_1,\cdots,x_d)\in U$,}
    \]
    where each $g_{ij}(\xbf, y_1,\cdots,y_j)$ is a polynomial in $(\xbf,y_1,\cdots,y_j)\in\RR^{d+j}$ of degree at most $\alpha$.
    Given a Pfaffian chain $(f_1,\dots,f_\ell)$ on $U$, a function $f:U\to\RR$ is called a \emph{Pfaffian function} with \emph{Pfaffian format} $(d,\ell,\alpha,\beta)$, if there is a polynomial $p$ of degree $\beta$ such that
    \[
        f(\xbf)=p(\xbf, f_1(\xbf),\cdots,f_\ell(\xbf)).
    \]
\end{definition}

Indeed, the smooth activations of interest in modern deep learning practice are Pfaffian.

\begin{proposition}\label{prop:SiLU-GELU-Pfaffian}
    All of the $C^\infty$ activations in \eqref{eq:SiLU}–\eqref{eq:Mish} are Pfaffian function on $\RR$, and their Pfaffian formats are $(1,1,2,2)$ for SiLU, $(1,2,4,2)$ for GELU, and $(1,6,5,2)$ for Mish.
\end{proposition}

We next recall the definition of Pdim for real-valued function classes.

\begin{definition}[Pseudo-dimension~\cite{anthony2009neural}]
    Let $\Fcal \subset \RR^{\Xcal}$ be a class of real-valued functions. We say that $x_1,\dots,x_n \in \Xcal$ together with $y_1,\dots,y_n \in \RR$ are \emph{pseudo-shattered} by $\Fcal$ if
    \(
        \big\{(\,\II\{f(x_i)>y_i\})_{i=1}^n : f\in\Fcal \big\} = \{0,1\}^n.
    \)
    The \emph{pseudo-dimension} of $\Fcal$, denoted as $\Pdim(\Fcal)$, is the largest $n$ for which there exist $x_1,\dots,x_n$ and $y_1,\dots,y_n$ that are pseudo-shattered by $\Fcal$. 
\end{definition}

We now present an upper bound on the Pdim of smooth DNNs with Pfaffian activations.

\begin{theorem}\label{thm: pseudo-dimension of Pfaffian network}
    Let $\sigma\in C^\infty(\RR)$ be an analytic activation with Pfaffian format $(1,\ell,\alpha,\beta)$. Then, for some suppressed universal constant, the Pdim of the smooth DNN satisfies
    \[
        \Pdim(\Fcal(d,1,D, W, \sigma))\ \lesssim_{\log}\ D^4W^6,
    \]
    where the suppressed constant depend on $d$, Pfaffian format of $\sigma$, and poly-logarithmic terms of $D$ and $W$.
\end{theorem}

\subsection{Approximation Error}\label{sec: Approximation Error}

As the smooth DNN becomes deeper or wider, its increased Pdim may incur larger statistical error, while the increased capacity can reduce approximation error by representing more complex targets. Thus, non-asymptotic approximation bounds for smooth DNNs are required to identify architectural scalings that balance these two errors.

Building on approximation theory for ReLU FNNs~\cite{yarotsky2017error, lu2021deep}, recent studies have extended approximation guarantees to smooth FNNs. In particular, \cite{zhang2024deep} showed that any ReLU FNN can be approximated arbitrarily well by FNNs for a broad class of smooth activations. More recently, \cite{yang2025deep} leveraged linear combinations of smooth activations to approximate monomials and derived non-asymptotic approximation error bounds for such networks. Following the strategy in~\cite{yang2025deep}, we establish non-asymptotic approximation error bounds for smooth DNNs when approximating Sobolev functions and hierarchical composition models. We begin by stating the analytic assumptions on the activation $\sigma$.

\begin{assumption}\label{assumption: two conditions on the smooth activation function}
    Suppose the activation $\sigma:\RR\to \RR$ satisfies, for some $\mfrak\geq 3$,
    \begin{itemize}
        \item (\(\mfrak\)-th order quasi-decay) Let $H(x):=\II(x>0)$ be the Heaviside step function, and  define $\psi(x):=\sigma(x)/x$. There exist constants $C,G>0$ such that, for any $x\neq 0$ and $0\leq k\leq \mfrak$,
        \[
            |\psi^{(k)}(x)-H^{(k)}(x)|\leq \min\Bigl\{\frac{C}{|x|^{k+1}},\, G\Bigr\}.
        \]

        \item (Local $C^\infty$) There exist $a\in\RR$ and $\delta_\ast>0$ such that
        \[
            \sigma\in C^\infty(a-\delta_\ast, a+\delta_\ast),\quad \sigma^\prime(a)\neq 0,\quad \text{and}\quad \sigma^{\pprime}(a)\neq 0.
        \]
    \end{itemize}    
\end{assumption}

This assumption ensures that smooth DNNs can approximate monomials and two specific functions that are central in neural network approximation theory. For the $C^\infty$ activations of interest in this work, including SiLU, GELU, and Mish, the local $C^\infty$ assumption is satisfied. Since these activations admit a uniformly bounded and increasing function $\psi(x)=\sigma(x)/x$ with exponentially decaying derivatives, the $\mfrak$-th order quasi-decay assumption also holds.

We now present the approximation error bounds of smooth DNNs for Sobolev functions.

\begin{theorem}\label{thm: approximation error and sobolev bound on Omega for sobolev smooth function}
    Suppose Assumption~\ref{assumption: two conditions on the smooth activation function} holds. Then, for any Sobolev function $f\in W^{\alpha,\infty}(\Omega)$ and any $W,D\in\NN_+$ satisfying $\log_2 W\leq D$, there exists a smooth DNN $\phi$ with width $\lesssim W\,\log(W)$ and depth $\lesssim D\,\log(D)$, where the suppressed constants depend only on $d$, $\alpha$, and $\mfrak$, such that, for $s=0,1,\cdots, \mfrak\land \lfloor\alpha\rfloor$,
    \begin{equation}\label{eq: first result in thm: approximation error and sobolev bound on Omega for sobolev smooth function}
        \|f-\phi\|_{W^{s,\infty}(\Omega)}\leq C_{16}(\alpha,d,\mfrak)\|f\|_{W^{\alpha,\infty}(\Omega)}\, (WD)^{-\frac{2(\alpha-s)}{d}}.
    \end{equation}
\end{theorem}

This result establishes approximation rates for smooth DNNs when approximating Sobolev functions. In what follows, we establish non-asymptotic approximation error bounds for smooth DNNs under hierarchical composition models. 

\begin{definition}[Hierarchical composition model]
    Given $d,\ell\in\NN$, $C>0$, and $\Pscr\subseteq \NN_+\times\NN_+$, such that $\sup_{(\beta,t)\in\Pscr}\beta\lor t<\infty$, the hierarchical composition model $\Hcal(d,\ell,\Pscr, C)$ is defined recursively as follows. For $\ell=1$, 
    $$\begin{aligned}
        \Hcal(d,1,\Pscr, C)
        =&\{h:\RR^d\to\RR:h(x)=g(x_{\pi(1)},\cdots,x_{\pi(t)}),\text{ where }\pi:[t]\to [d]\text{ and }\\
        &\qquad g:\RR^t\to\RR\text{ is }(\beta,C)\text{-smooth for some }(\beta,t)\in\Pscr\};
    \end{aligned}$$
    and for $\ell>1$, 
    $$\begin{aligned}
        \Hcal(d,\ell,\Pscr, C)
        =&\{h:\RR^d\to\RR:h(x)=g(f_1(x),\cdots,f_t(x)),\text{ where }f_i\in\Hcal(d,\ell-1,\Pscr)\text{ and }\\
        &\qquad g:\RR^t\to\RR\text{ is }(\beta,C)\text{-smooth for some }(\beta,t)\in\Pscr\}.
    \end{aligned}$$
    The adjusted model complexity of $\Hcal(d,\ell,\Pscr, C)$ is defined as $\gamma^\ast=\min_{(\beta,t)\in\Pscr}\beta/t$.
\end{definition}

Leveraging results in Theorem~\ref{thm: approximation error and sobolev bound on Omega for sobolev smooth function}, we obtain the following approximation error results for smooth DNNs under hierarchical composition models.

\begin{theorem}\label{thm: approximation error and holder bound on Omega for hierarchical composition model}
    For any $f_0\in \Hcal(d,\ell,\Pscr,C)$, $W,D\in\NN_+$ with $\log_2 W\leq D$, and $W\geq \max\{t:(\beta,t)\in\Pscr\}$, there is a $\sigma$-activated smooth DNN $\phi_{f_0}$ with width
    \(
        C_{19}(d, \mfrak, \ell, \Pscr)\cdot W\log(W),
    \)
   depth
    \(
        C_{19}(d, \mfrak, \ell, \Pscr)\cdot D\log(D),
    \)
    and activation $\sigma$ satisfying Assumption~\ref{assumption: two conditions on the smooth activation function}, such that 
    \begin{equation}\label{eq: first result in thm: approximation error and holder bound on Omega for hierarchical composition model}
        \|f_0-\phi_{f_0}\|_{L^\infty([0,1]^d)}\leq C_{18}(d,\mfrak,C,\ell, \Pscr)\, (WD)^{-2\gamma^\ast}.
    \end{equation}
\end{theorem}

Theorem~\ref{thm: approximation error and holder bound on Omega for hierarchical composition model} shows that when the unknown target $f_0$ admits a hierarchical composition of several $(\beta,C)$-smooth functions, each with either high smoothness or low input dimension, smooth DNNs can adapt to this intrinsic low-dimensional structure without explicit knowledge of the underlying composition. In particular, smooth DNNs achieve the same approximation bounds as ReLU FNNs under the hierarchical composition model~\cite{bauer2019deep, schmidt2020nonparametric, kohler2021rate}. 

\subsection{Upper Bounds of Hölder Norms}\label{sec: Upper Bounds of Hölder Norms}

As discussed in Section~\ref{sec: Origin of Uniform Convergence}, uniform convergence rates are governed by the smoothness of the candidate function class. Consequently, to establish uniform convergence rates for smooth DNN estimators, we need to characterize their Hölder norms across different orders. 

We first examine the analytic properties of the activations SiLU, GELU, and Mish.

\begin{proposition}\label{prop: the structure of Hölder norms of activations}
    For the activation function $\sigma$ corresponding to SiLU, GELU, and Mish, there exists a constant $C_\sigma>0$, depending only on $\sigma$, such that
    \[
        \sup_{z\in\RR}|\sigma^{(m)}(z)|\leq (C_\sigma\, m)^m,\quad\text{for all } m\geq 1.
    \]
\end{proposition}

Proposition~\ref{prop: the structure of Hölder norms of activations} shows that the derivatives of the smooth activations of interest grow at a factorial rate. As these activations are the building blocks of smooth DNNs, this result provides a natural baseline for the order of the Hölder-norm bounds that one can expect for smooth DNNs. The following result further establishes factorial-order Hölder norm bounds for smooth DNN approximators on a subset of $\Omega$ that is arbitrarily close to $\Omega$.

\begin{theorem}\label{thm: holder norm of DSNN approximator}
    Under the conditions of Theorem~\ref{thm: approximation error and sobolev bound on Omega for sobolev smooth function}, for any $\delta\in(0,1)$, there exists a smooth DNN $\phi$, whose activation $\sigma$ satisfies Assumption~\ref{assumption: two conditions on the smooth activation function}, with the same width and depth as in Theorem~\ref{thm: approximation error and sobolev bound on Omega for sobolev smooth function}, such that \eqref{eq: first result in thm: approximation error and sobolev bound on Omega for sobolev smooth function} continues to hold, and there exists a measurable subset $\Omega_{\delta}\subseteq\Omega$ with the Lebesgue measure $\lambda_d(\Omega_{\delta})\geq 1-\delta$ satisfying
    \begin{equation}\label{eq: result in thm: holder norm of DSNN approximator}
        \|\phi\|_{C^{\mfrak}(\Omega_{\delta})}\leq \Bigl(C_{19}(\alpha,d)\,\mfrak\Bigr)^\mfrak\,\|f\|_{W^{\alpha,\infty}(\Omega)}.
    \end{equation}
\end{theorem}

This result shows that the smooth DNN approximator admits a factorial-order upper bound on the Hölder norm $C^{\mfrak}$, as that for analytic activations in Proposition~\ref{prop: the structure of Hölder norms of activations}, even when $\mfrak>\alpha$ exceeds the smoothness order of the target function. Moreover, this result does not require $\sigma$ to be analytic, indicating that the factorial-order upper bound originates from the intrinsic structure of smooth DNNs rather than from the analyticity of $\sigma$. 

Rather than establishing a factorial-order upper bound on the entire domain $\Omega=[0,1]^d$, we obtain such a bound only on a subset $\Omega_{\delta}$, whose measure can be made arbitrarily close to that of $\Omega$. This restriction arises because the smooth DNN approximator $\phi$ involves certain sub-networks whose Sobolev norm control is difficult to establish on small regions of the domain. That said, we believe that the present result is sufficient to explain the well-behaved Hölder norms of smooth DNN approximators. 

\section{Uniform Convergence of Smooth DNN Estimators}\label{sec: Applications of DSNNs}

This section derives uniform convergence rates for smooth DNN estimators under several widely used models. Specifically, in Section~\ref{sec: Huber Regression}, we study smooth DNN estimators under the Huber regression model, establish their robustness to heavy-tailed noise, and consider least-squares regression as a special case of Huber regression. We then investigate quantile regression with smooth DNN estimators in Section~\ref{sec: Quantile Regression}. Uniform convergence rates for smooth DNN estimators under logistic regression are established in Section~\ref{sec: Generalized Linear Models}.

Before turning to these models, we assume the covariate $X_i$ is supported on $\Omega:=[0,1]^d$ and admits a bounded density throughout this section. In particular, there exist constants $0<c<C<\infty$, such as the density $p_X$ of $X_i$ satisfies 
\[
    0<c<p_X(x)\leq C<\infty,\quad\text{for all }x\in[0,1]^d.
\]
We also impose the following assumption on the activation function $\sigma$ to invoke the theoretical properties of smooth DNNs established in Section~\ref{sec: Statistical Properties of DSNNs}. SiLU, GELU, and Mish are key examples satisfying these conditions.

\begin{assumption}\label{assumption: activation function}
    The activation $\sigma$ is Pfaffian and satisfies Assumption~\ref{assumption: two conditions on the smooth activation function} for any $\mfrak\in\NN$. 
\end{assumption}

To leverage interpolation inequalities when deriving uniform convergence rates, we impose the following assumption on the Hölder norms of the candidate function class. As discussed in Section~\ref{sec: Upper Bounds of Hölder Norms}, the rationale is that both analytic activations and smooth DNN approximators to Sobolev functions satisfy factorial-type upper bounds on their Hölder norms. Thus, this condition can be interpreted as restricting attention to a bounded subset of smooth DNNs in a neighborhood of the approximator.

\begin{assumption}\label{assumption: DSNN class with holder norm control}
    For target function $f_0\in \Hcal(d,\ell,\Pscr, C)$, define the approximator
    \[
        f_n^\ast\in \argmin_{f\in \Fcal(d, 1, D, W, \sigma)}\|f-f_0\|_{L^\infty(\Omega)}. 
    \]
    We assume that $f_n^\ast\in \Fcal(d, 1, D, W, \sigma; A)$ for some $A>0$, where 
    \[
        \Fcal(d,1, D, W, \sigma; A) = \{f\in \Fcal(d, 1, D, W, \sigma): \|f\|_{C^\mfrak(\Omega)}\leq (A\, \mfrak)^\mfrak\text{ for all $\mfrak\in\NN$}\}.
    \]
\end{assumption}

\subsection{Huber Regression and Least Squares}\label{sec: Huber Regression}

In this section, we study the uniform convergence of smooth DNN estimators in Huber regression and least squares. Specifically, let $\{(X_i,Y_i)\}_{i=1}^n\sim P$ be \emph{i.i.d.}\ observations satisfying
\begin{equation}\label{eq: definition of regression data generating process}
    Y_i = f_0(X_i) + \xi_i, \qquad \text{where} \quad \EE[\xi_i \mid X_i] = 0,
\end{equation}
where $f_0$ denotes the target function of interest.

Motivated by the prevalence of heavy-tailed data in real-world applications, a growing line of work has investigated the $L^2$ estimation error for nonparametric function estimators under heavy-tailed noise~\cite{han2019convergence, kuchibhotla2022least, fan2024noise, ding2025new}. In contrast to light-tailed noise, heavy-tailed noise produces outliers in $Y_i$ more frequently, making classical least-squares estimators unreliable. To address this issue, the Huber loss was introduced as a robust alternative, defined as, for the Huber parameter $\tau\geq 0$, 
\(
    \ell_\tau(x) = \frac{1}{2}x^2\II(|x|\leq\tau) + (\tau|x| - \frac{1}{2}\tau^2)\II(|x|>\tau).
\)
Then, for a function class $\Fcal_n$, the Huber estimator is defined as
\begin{equation}\label{eq: definition of deep huber estimator}
    \hat{f}_n(\tau)\in\argmin_{f\in\Fcal_n}\frac{1}{n}\sum_{i=1}^n \ell_\tau\big(Y_i-f(X_i)\big).
\end{equation}

Recent advances have characterized the robustness of the Huber estimator $\hat{f}_n(\tau)$ for different function classes with non-asymptotic analysis. When $\Fcal_n$ consists of linear functions, \cite{sun2020adaptive} showed that the adaptive Huber estimator achieves a sub-Gaussian concentration bound in the presence of heavy-tailed noise. More recently, analogous robustness guarantees for ReLU FNN estimators were established in prior works~\cite{fan2024noise, ding2025new}. However, these results are formulated in the $L^2(P)$ norm rather than in $L^\infty$, leaving open whether DNN estimators remain robust in the uniform norm under heavy-tailed regimes. 

The assumption below is imposed to show that smooth DNN Huber estimators are robust in the uniform norm under heavy-tailed noise.

\begin{assumption}\label{assumption: distributional assumption for Huber regression}
    For the model in \eqref{eq: definition of regression data generating process}, assume:
    \begin{itemize}
        \item \textbf{Target function:} For some $M>1$, $\|f_0\|_{L^\infty([0,1]^d)}\leq M$.
        \item \textbf{Noise:} For some $m>1$, there exists $v_m\in(0,\infty)$ such that $\|\EE[|\xi_i|^m\mid X_i]\|_{L^\infty}\leq v_m$.
    \end{itemize}
\end{assumption}

We now establish robustness guarantees for the estimator $\hat{f}_n(\tau)$ in the $L^\infty(\Omega)$ norm.

\begin{theorem}\label{thm: uniform convergence rate of deep smooth Huber regression}
    Suppose Assumptions~\ref{assumption: activation function},~\ref{assumption: DSNN class with holder norm control} and~\ref{assumption: distributional assumption for Huber regression} (taking $v_2=\infty$ if $m<2$) hold, and assume 
    \(
        \tau\geq 2\max\{2M,(2v_m)^{1/m}\}.
    \)
    Let $\Fcal_n=\Fcal(d,1,D,W,\sigma;A)\cap L^\infty(M)$ be the uniformly bounded smooth DNN class.
    Denote the effective sample size as $\tilde{n}=n/(D^4W^6)$ and assume $\tilde{n}\geq1$. Then, for any $\delta\in(0,1)$, with probability at least $1-\delta$,
    \[\begin{aligned}
        &\|\hat{f}_n(\tau)-f_0\|_{L^\infty(\Omega)}
        \lesssim_{\log} \Bigl(\delta_\sfrak +(WD)^{-2\gamma^\ast} + \frac{v_m}{\tau^{m-1}} + \sqrt{\tau\land (\sqrt{v_2}+ M)}\, \sqrt{\frac{\tau\,\log(10/\delta)}{n}}\Bigr)^{0.99}.\\
    \end{aligned}\]
    where the exponent $0.99$ can be replaced by any constant in $(0,1)$. The suppressed constant depends on $d$, $A$, $C$, $\ell$, $\Pscr$, $\sigma$, $P$, and polylogarithmic terms in $n$, $W$, and $D$. Moreover, 
    \begin{equation}\label{eq: statistical error in uniform convergence rate of deep smooth Huber regression}
        \delta_\sfrak=\begin{cases}
            \sqrt{(M +\sqrt{v_2})(M+v_m^{1/m})}\cdot \tilde{n}^{1/(2m)-1/2}  & \text{if $\tau\geq \tilde{n}^{\frac{1}{m}}\,(M+v_m^{1/m})$ and $m\geq 2$,}\\
            \sqrt{\tau\land (M+\sqrt{v_2})}\cdot \sqrt{\tau/\tilde{n}} &  \text{otherwise.}
        \end{cases}
    \end{equation}
\end{theorem}

In this result, the uniform estimation error of the smooth DNN Huber estimator $\hat{f}_n(\tau)$ is decomposed into four components: (i) statistical error $\delta_\sfrak$, (ii) approximation error $(WD)^{-2\gamma^\ast}$, (iii) Huberization bias $v_m/\tau^{m-1}$, and (iv) a sub-Gaussian deviation term. These components reveal two natural trade-offs. First, increasing $\tau$ reduces the Huber bias but enlarges both the deviation term and $\delta_\sfrak$, since the Huber loss approaches the least-squares loss. Second, enlarging the network architecture (in terms of $D$ and $W$) decreases the approximation error but reduces the effective sample size $\tilde{n}$, thereby increasing the statistical error. Moreover, this $L^\infty$ estimation error decomposition parallels that of ReLU Huber estimators in the $L^2$ norm~\cite{fan2024noise,ding2025new}, with the only distinction arising from the pseudo-dimensions of the two function classes. This comparison highlights a concrete advantage of smooth DNN estimators over existing ReLU FNNs in terms of uniform convergence guarantees.

We now turn to least-squares regression as a special case of Huber regression. As shown in Section~\ref{sec: Limitations of ReLU Networks in Uniform Convergence}, ReLU FNN least-squares estimators suffer from the \textit{curse of dimensionality} even under idealized settings, in sharp contrast to their favorable $L^2$ convergence behavior established in~\cite{fan2024noise,ding2025new}. Since the least-squares loss can be viewed as the limiting case of the Huber loss $\ell_\tau$ as $\tau\to\infty$, Theorem~\ref{thm: uniform convergence rate of deep smooth Huber regression} suggests the corresponding uniform convergence behavior of smooth DNN least-squares estimators under heavy-tailed noise. However, one cannot directly let $\tau\to\infty$ in Theorem~\ref{thm: uniform convergence rate of deep smooth Huber regression}, because the final term, namely the sub-Gaussian deviation bound, diverges in this limit. This is precisely because least-squares regression is not robust to heavy-tailed noise, and therefore cannot satisfy a non-asymptotic heavy-tailed robustness guarantee of sub-Gaussian type. For this reason, in Theorem~\ref{theorem: uniform convergence rate of deep smooth LS regression} below, we formulate the uniform estimation error of smooth DNN least-squares estimators in terms of convergence in expectation rather than a non-asymptotic deviation bound.

\begin{theorem}\label{theorem: uniform convergence rate of deep smooth LS regression}
    Under the conditions and notation of Theorem~\ref{thm: uniform convergence rate of deep smooth Huber regression}, when $m\ge 2$, the smooth DNN least-squares estimator $\hat f_n$ satisfies
    \[
        \EE\|\hat{f}_n-f_0\|_{L^\infty(\Omega)} \lesssim_{\log} \Bigl((M+v_m^{1/m})\cdot\tilde{n}^{1/(2m)-1/2} +(WD)^{-2\gamma^\ast} \Bigr)^{0.99}.\\
    \]
    In particular, minimizing over $D$ and $W$ yields
    \[
        \EE\|\hat{f}_n-f_0\|_{L^\infty(\Omega)} \lesssim_{\log} n^{-\frac{0.99\, \gamma^\ast (1-1/m)}{2\gamma^\ast + 2(1-1/m)}}.\\
    \]
\end{theorem}

While Theorem~\ref{thm: lower bound of ReLU with penalty} shows that the convergence rate of ReLU FNN least-squares estimators can be as slow as $n^{-\frac{1}{d+1}}$ regardless of the smoothness of $f_0$, smooth DNN least-squares estimators can adapt to the hierarchical composition structure of $f_0$, so that their convergence rates depend on the intrinsic dimensionality of its low-dimensional representation. When the noise is light-tailed, as in Theorem~\ref{thm: lower bound of ReLU with penalty} (in the sense that $m\to\infty$), the resulting rate in Theorem~\ref{theorem: uniform convergence rate of deep smooth LS regression} scales as $n^{-\frac{0.99\,\gamma^\ast}{2\gamma^\ast+2}}$. Thus, smooth DNN estimators can mitigate the \textit{curse of dimensionality} for least-squares regression in the uniform norm by adapting to the low-dimensional hierarchical composition structure of $f_0$.

\subsection{Quantile Regression}\label{sec: Quantile Regression}

As another robust estimation task, quantile regression estimates conditional quantile functions~\cite{koenker2005quantile}. For a fixed quantile level $\tau \in (0,1)$, suppose we observe \emph{i.i.d.}\ samples $\{(X_i, Y_i)\}_{i=1}^n$ such that the $\tau$-th conditional quantile of $Y_i$ given $X_i = x$ is $f_0(x)$. Then $f_0$ minimizes the population quantile regression loss: for the check loss $\rho_\tau(u) = u(\tau - \II(u < 0))$,
\[
    f_0 \in \argmin \EE[\rho_\tau(Y_1 - f(X_1))].
\]
Recent work has established non-asymptotic $L^2(P)$ convergence rates and non-asymptotic robustness guarantees for ReLU FNN quantile regression estimators~\cite{ding2025new, yu2025deep}.
In the following, we establish robustness guarantees for smooth DNN quantile estimators in the $L^\infty$ norm. Specifically, we introduce the following regularity condition to ensure well-behaved conditional densities in a neighborhood of the target quantile: 

\begin{assumption}\label{assumption: adaptive self-calibration condition of conditional distribution}
    Assume for some $M>1$, $\|f_0\|_{L^\infty([0,1]^d)}\leq M$. Denote $p_{Y|X=x}$ as the condition density of $Y$ given $X=x$. Assume there exists some $\delta>0$, such that $0<\inf_{t\in[f_0(x)-\delta,f_0(x)+\delta]}p_{Y|X=x}(t)\leq \sup_{t\in\RR}p_{Y|X=x}(t)<\infty,$ almost surely.
\end{assumption}

Leveraging the theoretical results developed in Section~\ref{sec: Statistical Properties of DSNNs}, we establish robustness guarantees for the smooth DNN quantile regression estimator in the $L^\infty$ norm below.

\begin{theorem}\label{thm: uniform convergence rate for DSNN quantile regression}
    Suppose Assumptions~\ref{assumption: activation function},~\ref{assumption: DSNN class with holder norm control} and~\ref{assumption: adaptive self-calibration condition of conditional distribution} holds, and denote the smooth DNN class 
    \(
        \Fcal_n=\Fcal(d,1,D,W,\sigma;A)\cap L^\infty(M).
    \)
    Define the smooth DNN quantile estimator as
    \[
        \hat{f}_n\in\argmin_{f\in\Fcal_n}\sum_{i=1}^n \rho_\tau(Y_i-f(X_i)).
    \]
    Denote the effective sample size as $\tilde{n}=n/(D^4W^6)$. Then, for any $\delta\in(0,1)$, 
    \[
        \PP\Bigl(\|\hat{f}_n-f_0\|_{L^\infty(\Omega)}\gtrsim_{\log} \bigl(\sqrt{\log(2/\delta)}\cdot\tilde{n}^{-\frac{1}{2}} + (WD)^{-2\gamma^\ast}\bigr)^{0.99}\Bigr)\leq \delta.
    \]
\end{theorem}

Similar to Theorem~\ref{thm: uniform convergence rate of deep smooth Huber regression}, the exponent $0.99$ is flexible and can be replaced by any value strictly smaller than $1$. Moreover, by optimizing over the depth and width of smooth DNNs in terms of $(D,W)$, the convergence rate of the smooth DNN quantile estimator scales as
\(
    \EE\|\hat{f}_n-f_0\|_{L^\infty(\Omega)}\lesssim_{\log} n^{-\frac{0.99\,\gamma^\ast}{2+2\gamma^\ast}}.
\)
Notably, this convergence exponent does not depend on the ambient dimension $d$, but only on the adjusted model-complexity parameter $\gamma^\ast$. This justifies that the smooth DNN estimator can mitigate the \textit{curse of dimensionality} by adapting to the low-dimensional hierarchical composition structure of $f_0$.

\subsection{Logistic Regression}\label{sec: Generalized Linear Models}

In this subsection, we first establish uniform estimation error bounds for smooth DNN logistic regression estimators in Theorem~\ref{theorem: uniform convergence rate for DSNN logistic regression}. Building on this result, Theorem~\ref{theorem: uniform convergence rate for density-ratio based distribution estimate} further establishes uniform convergence guarantees for the density-ratio-based distribution estimator studied in \cite{manole2024background}. For \emph{i.i.d.}\ observations $\{(X_i,Y_i)\}_{i=1}^n$, define the smooth DNN logistic regression estimator by
\[
    \hat{f}_n\in \argmin_{f\in\Fcal_n}\frac{1}{n}\sum_{i=1}^n -Y_i\cdot f(X_i)+\log(1 + \exp(f(X_i)))=:\frac{1}{n}\sum_{i=1}^n\ell(f;X_i,y_i),
\]
where $\Fcal_n = \Fcal(d,1,D,W,\sigma;A)\cap L^\infty(M)$. Let the target function $f_0$ satisfy
\[
    \EE[Y_i|X_i]= \frac{\exp(f_0(X_i))}{1 + \exp(f_0(X_i))}.
\]

As in the cases of Huber and quantile regression, convergence rates for the smooth DNN logistic regression estimator can be established below by leveraging the theoretical results developed in Section~\ref{sec: Statistical Properties of DSNNs}.
\begin{theorem}\label{theorem: uniform convergence rate for DSNN logistic regression}
    Suppose Assumptions~\ref{assumption: activation function},~\ref{assumption: DSNN class with holder norm control} hold. Assume $\|f_0\|_{L^\infty(\Omega)}\leq M$. Denote the effective sample size as $\tilde{n}=n/(D^4W^6)$. Then, for any $\delta\in(0,1)$, 
    \[
        \PP\Bigl(\|\hat{f}_n-f_0\|_{L^\infty(\Omega)}\gtrsim_{\log} \bigl(\sqrt{\frac{\log(10/\delta)}{\tilde{n}}} + (WD)^{-2\gamma^\ast}\bigr)^{0.99}\Bigr)\leq \delta.
    \]
    where the exponent $0.99$ can be replaced by any constant in $(0,1)$.
\end{theorem}

Theorem~\ref{theorem: uniform convergence rate for DSNN logistic regression} shows that the smooth DNN logistic regression estimator enjoys an estimation error bound analogous to that of the smooth DNN quantile regression estimator established in Theorem~\ref{thm: uniform convergence rate for DSNN quantile regression}. This further supports the reliability of smooth DNN estimators across a range of regression and classification models.

The uniform convergence guarantee established in Theorem~\ref{theorem: uniform convergence rate for DSNN logistic regression} is also useful for supporting the practical application of smooth DNNs to estimating the probability of Higgs boson pair production decaying into four bottom quarks (denoted by $HH\to 4b$)~\cite{manole2024background}. We briefly reformulate this problem within the nonparametric framework considered in \cite{manole2024background} and then state the corresponding uniform convergence guarantee.

In this application, the event $HH\to 4b$ is the signal of interest, while other physical processes producing four bottom quarks form the background. Accurate estimation of both signal and background distributions is needed to distinguish signal from background and conduct subsequent hypothesis testing. While the signal distribution can be approximated using physical simulation, simulating the background distribution is computationally prohibitive. Consequently, the main problem in \cite{manole2024background} is to estimate the background distribution from observed data contaminated by signal events.

To this end, \cite{manole2024background} introduced an auxiliary distribution together with density ratio estimation. Let $\Omega\subseteq\RR^d$ denote the state space of the observed physical quantities, and let $P$ and $Q$ denote the probability measures of the background and auxiliary distributions on $\Omega$, respectively. In that work, $Q$ corresponds to events with three observed bottom quarks, which share similar physical characteristics with background events. Density ratio estimation is reduced to a binary classification problem: let $Y$ be binary and let $E$ satisfy
\[
    E\mid Y=0\sim Q,\qquad E\mid Y=1\sim P.
\]
Define
\[
    \psi(x):=\PP(Y=1\mid E=x),\qquad x\in\Omega.
\]
Then Bayes' rule gives, when $\PP(Y=1)=\PP(Y=0)=0.5$,
\[
    P(A)=\int_A \frac{\psi(y)}{1-\psi(y)}\,Q(\dd y),\qquad \text{for any measurable }A\subseteq\Omega.
\]
Hence, if $\hat\psi$ is an estimator of $\psi$ and $\QQ_N$ is an empirical measure of $Q$, then a plug-in estimator of $P$ is
\[
    \hat P(A)=\int_A \frac{\hat\psi(y)}{1-\hat\psi(y)}\,\QQ_N(\dd y),\qquad \text{for any measurable }A\subseteq\Omega.
\]
Similarly, if $p$ and $q$ are the densities of $P$ and $Q$, and $\hat q$ is an estimator of $q$, then
\[
    \hat p(x)=\frac{\hat\psi(x)}{1-\hat\psi(x)}\,\hat q(x),\qquad x\in\Omega.
\]
In \cite{manole2024background}, the density ratio $\frac{\psi(y)}{1-\psi(y)}$ is estimated using a SiLU ResNet tailored to the structure and symmetries of the collider data. 

A key difficulty is that the observed four-bottom-quark data are contaminated by signal events, so one cannot directly fit a logistic regression model on the full domain. Following \cite{manole2024background}, suppose
\[
    \Omega=\Xcal\cup\Ycal,\qquad \Xcal\cap\Ycal=\emptyset,
\]
where signal events occur only on $\Xcal$. Then contamination is present only on $\Xcal$, while the observed four-bottom quark events on $\Ycal$ are purely background. Therefore, one may estimate the classifier between background and three-bottom quark events in $\Ycal$, and then extrapolate it to the full domain $\Omega$. Although this extrapolation assumption is statistically strong, it is reasonable in this physical setting and can be cross-checked against an alternative optimal-transport-based approach. To establish theoretical guarantees for \cite{manole2024background}, we adopt this framework in a slightly relaxed form.

\begin{assumption}\label{assumption: safe expolation for manole2024background}
    Let $\widetilde{P}$ denote the probability measure induced by the combined distribution of $P$ and $Q$, and let $\widetilde{P}_\Ycal$ denote the restriction of $\widetilde{P}$ to $\Ycal$. For the candidate smooth DNN class $\Fcal_n$, assume that there exists $\varepsilon_n>0$ such that
    \(
        \sup_{f\in\Fcal_n}|\widetilde{P}\ell(f) - \widetilde{P}_\Ycal\ell(f)|\leq \varepsilon_n.
    \)
\end{assumption}
Compared with Assumption 1 in \cite{manole2024background}, which requires the binary classifier on $\Ycal$ to coincide exactly with the binary classifier on the whole domain $\Omega$, the above assumption allows the two population risk minimizers to differ and quantifies the resulting discrepancy between the corresponding population loss functions through $\varepsilon_n$. In particular, when $\widetilde{P}$ and $\widetilde{P}_\Ycal$ exhibit similar physical patterns for distinguishing four-bottom-quark events from three-bottom-quark events, one expects $\varepsilon_n$ to be small.

Leveraging Assumption~\ref{assumption: safe expolation for manole2024background} together with the uniform convergence guarantee in Theorem~\ref{theorem: uniform convergence rate for DSNN logistic regression}, we establish the following result for the density-ratio-based distribution estimators.

\begin{theorem}\label{theorem: uniform convergence rate for density-ratio based distribution estimate}
    Under the conditions of Theorem~\ref{theorem: uniform convergence rate for DSNN logistic regression}, suppose in addition that Assumption~\ref{assumption: safe expolation for manole2024background} holds. Let $X_1,\ldots,X_n$ be i.i.d.\ observations from $\widetilde{P}_\Ycal$, and let $Y_i=1$ if $X_i$ corresponds to a four-bottom-quark background event and $Y_i=0$ if $X_i$ corresponds to a three-bottom-quark auxiliary event. Let $\hat f_n$ be the smooth DNN logistic regression estimator based on $\{(X_i,Y_i)\}_{i=1}^n$, and define the plug-in estimator of $\psi$ by
    \[
        \hat\psi(x)=\frac{\exp(\hat f_n(x))}{1+\exp(\hat f_n(x))}.
    \]
    Then, for any $\delta\in(0,1)$, the following bounds hold with probability at least $1-\delta$:
    \[
        |\hat{P}(A) - P(A)|\lesssim_{\log} \Bigl(\sqrt{\frac{\log(10/\delta)}{\tilde{n}}} + (WD)^{-2\gamma^\ast} + \sqrt{\varepsilon_n}\Bigr)^{0.99} + \Bigl|(Q-\QQ_N)\exp(f_0(\cdot))\II(A)\Bigr|
    \]
    for any measurable $A$, and 
    \[
        \|\hat{p}-p\|_{L^\infty(\Omega)}\lesssim_{\log} \Bigl(\sqrt{\frac{\log(10/\delta)}{\tilde{n}}} + (WD)^{-2\gamma^\ast} + \sqrt{\varepsilon_n}\Bigr)^{0.99} +   \|\hat q - q\|_{L^\infty(\Omega)}.
    \]
    Here, the exponent $0.99$ may be replaced by any constant in $(0,1)$.
\end{theorem}

Theorem~\ref{theorem: uniform convergence rate for density-ratio based distribution estimate} transfers the uniform convergence rate of the smooth DNN logistic regression estimator to uniform convergence guarantees for the density-ratio-based estimators $\hat{P}$ and $\hat{p}$ of the background distribution. This provides a theoretical justification for the density-ratio-based method in \cite{manole2024background} under the relaxed extrapolation condition in Assumption~\ref{assumption: safe expolation for manole2024background}.

In particular, the theorem yields non-asymptotic estimation error bounds for both $\hat P$ and $\hat p$. Each bound consists of four components: the statistical error term $\sqrt{\log(10/\delta)/\tilde{n}}$, the approximation error term $(WD)^{-2\gamma^\ast}$, the extrapolation error $\varepsilon_n$, and the estimation error of the auxiliary distribution, represented by $|(Q-\QQ_N)\exp(f_0(\cdot))\II(A)|$ and $\|\hat q-q\|_{L^\infty(\Omega)}$, respectively. The first two terms arise from smooth DNN logistic regression, while the term $\varepsilon_n$ quantifies the discrepancy introduced by extrapolation, which is expected to be small in this physical setting. Moreover, when the auxiliary distribution can be simulated, the last term is typically negligible in this setting. Besides, if a collection of measurable sets $\Acal$ is sufficiently regular so that
\[
    \sup_{A\in\Acal}\Bigl|(Q-\QQ_N)\exp(f_0(\cdot))\II(A)\Bigr|\to 0\qquad\text{as }N\to\infty,
\]
and if $\|\hat q-q\|_{L^\infty(\Omega)}\to 0$ (e.g., permitted by kernel density estimation~\cite{jiang2017uniform}), the auxiliary-distribution estimation error can be made asymptotically negligible for sufficiently large $N$.

\section{Numerical Experiments}\label{sec: Numerical Studies}

Section~\ref{sec: Simulation Experiments} benchmarks SiLU FNN and ResNet Huber estimators against ReLU FNN estimators through simulation studies. In Section~\ref{sec:realdata_ozone}, we apply SiLU ResNet quantile estimators to analyze the temperature sensitivity of ozone concentrations across regions. All experiments were conducted on a computer running {Ubuntu 24.04.2 LTS} with an {AMD EPYC 7532 CPU}, {512 GB RAM}, and an {NVIDIA RTX 3090 GPU}.\footnote{The Python code used to reproduce both the simulation and real-data experiments is available in an anonymized GitHub repository at \url{https://anonymous.4open.science/r/Uniform_convergence-DF22/}.}

\subsection{Simulation Studies}\label{sec: Simulation Experiments}

In this subsection, we compare the numerical performance of SiLU DNN and ReLU FNN estimators for Huber regression. We consider \emph{i.i.d.}\ samples $\{(X_i,Y_i)\}_{i=1}^n$ generated from
\(
    Y_i = f_0(X_i) + g(X_i)\,\varepsilon_i,
\)
where the covariate $X_i\in\RR^8$ is sampled uniformly from $[0,1]^8$, and $\varepsilon_i$ is drawn from a Student's $t$ distribution with either 2 or 4 degrees of freedom ($t_2$ or $t_4$). Following~\cite{yu2025deep} to induce noise, the target and scale functions are defined as
\[\begin{aligned}
    f_0(x)=&\cos(2\pi x_1)+\frac{1}{1+\exp(-x_2-x_3)}+\frac{1}{(1+x_4+x_5)^3}+\frac{1}{x_6+\exp(x_7x_8)},\\
    g(x)=&\sin\!\Big(\frac{\pi(x_1+x_2)}{2}\Big)+\log\!\big(1+x_3^2x_4^2x_5^2\big)+\frac{x_8}{1+\exp(-x_6-x_7)}.
\end{aligned}\]

For the Huber estimator in~\eqref{eq: definition of deep huber estimator}, we set the Huber parameter to $\tau\in\{0.25, 0.5, 1, 2, 4\}$. The depth and width of the ReLU FNNs, SiLU FNNs, and SiLU ResNets are all set to $4$. For the SiLU ResNets, we take $\Lcal_1$ and $\Lcal_3$ in Definition~\ref{def: Residual block} to be identity maps so that all architectures have matched parameter sizes. 

\begin{table}[H]
\centering
\caption{$L^2(P)$ and $L^\infty([0,1]^8)$ estimation errors of the Huber estimators under $t_2$ noise, with estimated convergence exponents and bootstrap standard deviations (shown in parentheses).} 
\label{tab:huber-t2}
    \scalebox{0.78}{
\begin{tabular}{cc|cc|cc|cc|cc}
\hline
& & \multicolumn{2}{c|}{$n=512$} & \multicolumn{2}{c|}{$n=1024$} & \multicolumn{2}{c|}{$n=2048$} & \multicolumn{2}{c}{Estimated Convergence Exponents} \\ \hline
$\tau$ & Model & $L^2$ & $L^\infty$ & $L^2$ & $L^\infty$ & $L^2$ & $L^\infty$ & $L^2$ & $L^\infty$ \\ \hline
\multirow{3}{*}{0.25}
& ReLU FNN & 0.6938 & 2.2463 & 0.4881 & 1.9525 & 0.3515 & 1.6019 & 0.4905 (0.0473) & 0.2439 (0.0424) \\
& SiLU FNN & 0.7256 & \textbf{1.9504} & 0.5658 & 1.8361 & 0.2910 & 1.3285 & 0.6591 (0.0424) & 0.2770 (0.0400) \\
& SiLU ResNet & \textbf{0.6218} & 2.1807 & \textbf{0.4227} & \textbf{1.6011} & \textbf{0.2406} & \textbf{1.0836} & 0.6848 (0.0378) & 0.5045 (0.0333) \\ \hline
\multirow{3}{*}{0.5}
& ReLU FNN & 0.7001 & 2.2730 & 0.4920 & 1.9812 & 0.3465 & 1.6172 & 0.5073 (0.0472) & 0.2456 (0.0437) \\
& SiLU FNN & 0.7238 & \textbf{1.9469} & 0.5628 & 1.8129 & 0.2911 & 1.3084 & 0.6571 (0.0438) & 0.2867 (0.0353) \\
& SiLU ResNet & \textbf{0.6171} & 2.1639 & \textbf{0.4199} & \textbf{1.5851} & \textbf{0.2399} & \textbf{1.0759} & 0.6815 (0.0395) & 0.5040 (0.0344) \\ \hline
\multirow{3}{*}{1.0}
& ReLU FNN & 0.6984 & 2.2195 & 0.4852 & 1.9804 & 0.3459 & 1.6315 & 0.5070 (0.0465) & 0.2220 (0.0427) \\
& SiLU FNN & 0.7236 & \textbf{1.9291} & 0.5688 & 1.8345 & 0.2942 & 1.3030 & 0.6491 (0.0439) & 0.2830 (0.0367) \\
& SiLU ResNet & \textbf{0.6128} & 2.1517 & \textbf{0.4146} & \textbf{1.5728} & \textbf{0.2406} & \textbf{1.1073} & 0.6743 (0.0394) & 0.4792 (0.0349) \\ \hline
\multirow{3}{*}{2.0}
& ReLU FNN & 0.7170 & 2.2358 & 0.5181 & 2.0449 & 0.3509 & 1.7148 & 0.5154 (0.0423) & 0.1914 (0.0461) \\
& SiLU FNN & 0.7317 & \textbf{1.9366} & 0.5985 & 1.8890 & 0.3210 & 1.3810 & 0.5944 (0.0420) & 0.2439 (0.0346) \\
& SiLU ResNet & \textbf{0.6355} & 2.2053 & \textbf{0.4497} & \textbf{1.6371} & \textbf{0.2512} & \textbf{1.1518} & 0.6694 (0.0337) & 0.4686 (0.0352) \\ \hline
\multirow{3}{*}{4.0}
& ReLU FNN & 0.7423 & 2.4542 & 0.5744 & 2.4412 & 0.3951 & 1.8325 & 0.4549 (0.0366) & 0.2107 (0.0575) \\
& SiLU FNN & 0.7489 & \textbf{2.0955} & 0.6335 & 2.0112 & 0.3458 & 1.5345 & 0.5573 (0.0371) & 0.2248 (0.0367) \\
& SiLU ResNet & \textbf{0.6746} & 2.4125 & \textbf{0.5025} & \textbf{1.7940} & \textbf{0.3024} & \textbf{1.3392} & 0.5789 (0.0338) & 0.4246 (0.0377) \\ \hline
\end{tabular}
}
\end{table}

\begin{table}[H]
\centering
\caption{$L^2(P)$ and $L^\infty([0,1]^8)$ estimation errors of the Huber estimators under $t_4$ noise, with estimated convergence exponents and bootstrap standard deviations (shown in parentheses).} 
\label{tab:huber-t4}
    \scalebox{0.78}{
\begin{tabular}{cc|cc|cc|cc|cc}
\hline
& & \multicolumn{2}{c|}{$n=512$} & \multicolumn{2}{c|}{$n=1024$} & \multicolumn{2}{c|}{$n=2048$} & \multicolumn{2}{c}{Estimated Convergence Exponents} \\ \hline
$\tau$ & Model & $L^2$ & $L^\infty$ & $L^2$ & $L^\infty$ & $L^2$ & $L^\infty$ & $L^2$ & $L^\infty$ \\ \hline
\multirow{3}{*}{0.25}
& ReLU FNN & 0.6990 & 2.1502 & 0.4797 & 1.7700 & 0.2942 & 1.4206 & 0.6242 (0.0477) & 0.2990 (0.0380) \\
& SiLU FNN & 0.7264 & \textbf{1.9761} & 0.5481 & 1.7115 & 0.2795 & 1.1868 & 0.6891 (0.0453) & 0.3678 (0.0293) \\
& SiLU ResNet & \textbf{0.6242} & 2.0273 & \textbf{0.3936} & \textbf{1.4715} & \textbf{0.2206} & \textbf{1.0343} & 0.7502 (0.0355) & 0.4854 (0.0271) \\ \hline
\multirow{3}{*}{0.5}
& ReLU FNN & 0.6922 & 2.1036 & 0.4719 & 1.8029 & 0.2930 & 1.4562 & 0.6202 (0.0497) & 0.2653 (0.0384) \\
& SiLU FNN & 0.7238 & \textbf{1.9601} & 0.5424 & 1.6897 & 0.2718 & 1.2117 & 0.7066 (0.0448) & 0.3469 (0.0362) \\
& SiLU ResNet & \textbf{0.6258} & 2.0186 & \textbf{0.3862} & \textbf{1.4541} & \textbf{0.2173} & \textbf{1.0177} & 0.7632 (0.0358) & 0.4941 (0.0279) \\ \hline
\multirow{3}{*}{1.0}
& ReLU FNN & 0.6921 & 2.0936 & 0.4730 & 1.7885 & 0.2888 & 1.5078 & 0.6304 (0.0486) & 0.2368 (0.0416) \\
& SiLU FNN & 0.7226 & \textbf{1.9242} & 0.5483 & 1.6163 & 0.2584 & 1.1810 & 0.7418 (0.0414) & 0.3521 (0.0352) \\
& SiLU ResNet & \textbf{0.6248} & 2.0070 & \textbf{0.3882} & \textbf{1.4337} & \textbf{0.2166} & \textbf{1.0055} & 0.7642 (0.0358) & 0.4985 (0.0290) \\ \hline
\multirow{3}{*}{2.0}
& ReLU FNN & 0.7074 & 2.0976 & 0.4810 & 1.7926 & 0.3182 & 1.5894 & 0.5763 (0.0487) & 0.2001 (0.0417) \\
& SiLU FNN & 0.7252 & \textbf{1.8967} & 0.5676 & 1.6364 & 0.2687 & 1.1916 & 0.7161 (0.0398) & 0.3353 (0.0289) \\
& SiLU ResNet & \textbf{0.6301} & 1.9698 & \textbf{0.4115} & \textbf{1.4732} & \textbf{0.2322} & \textbf{1.0322} & 0.7202 (0.0387) & 0.4662 (0.0252) \\ \hline
\multirow{3}{*}{4.0}
& ReLU FNN & 0.7236 & 2.1293 & 0.5236 & 1.8636 & 0.3341 & 1.6546 & 0.5575 (0.0455) & 0.1819 (0.0413) \\
& SiLU FNN & 0.7317 & \textbf{1.9123} & 0.5894 & 1.7204 & 0.2914 & 1.2531 & 0.6642 (0.0386) & 0.3049 (0.0250) \\
& SiLU ResNet & \textbf{0.6351} & 2.0273 & \textbf{0.4398} & \textbf{1.6076} & \textbf{0.2604} & \textbf{1.1379} & 0.6433 (0.0397) & 0.4166 (0.0256) \\ \hline
\end{tabular}
}
\end{table}

\begin{figure}[!htbp]
    \centering
    \includegraphics[width=\linewidth]{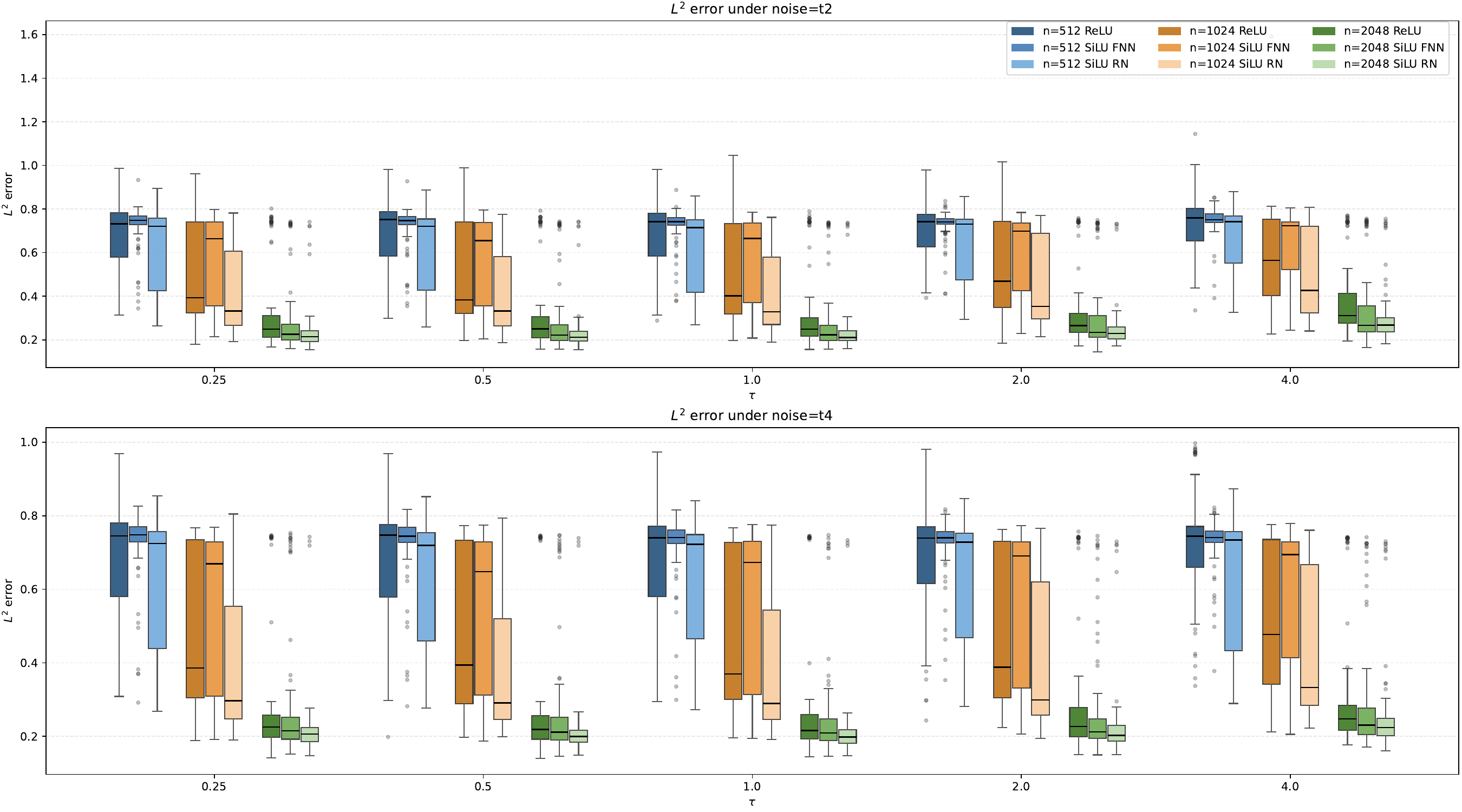}
    \caption{Side-by-side boxplots of the $L^2(P)$ estimation errors.}
    \label{fig:boxplot of loss, L2}
\end{figure}

\begin{figure}[!htbp]
    \centering
    \includegraphics[width=\linewidth]{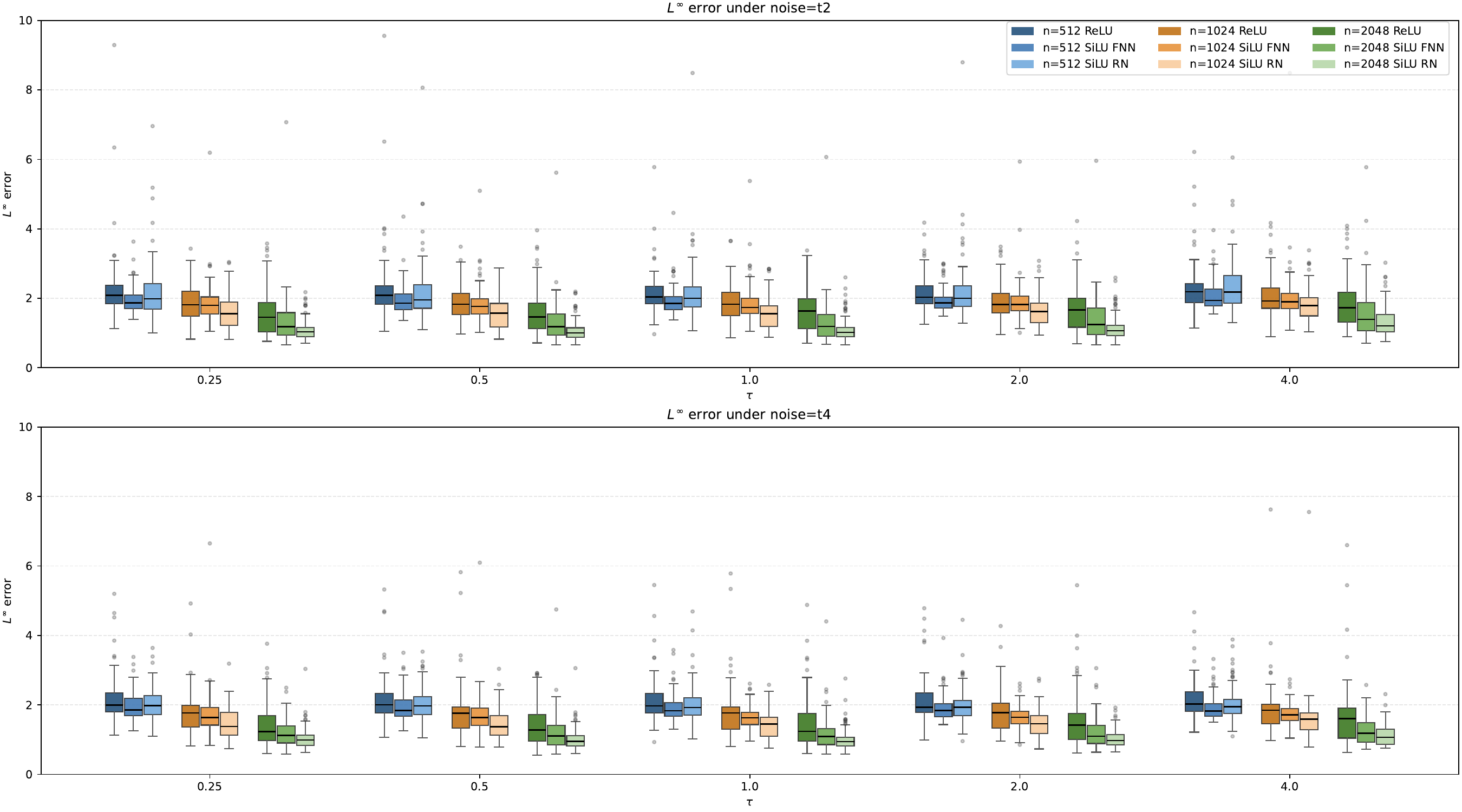}
    \caption{Side-by-side boxplots of the $L^\infty$ estimation errors.}
    \label{fig:boxplot of loss, Linf}
\end{figure}

We consider sample sizes $n\in\{512, 1024, 2048\}$ and replicate each setting 100 times. For evaluation, given an estimator $\hat{f}_n$, we independently draw $\{X_j\}_{j=1}^{10000}$ uniformly from $[0,1]^8$ and compute the pointwise errors $\{|\hat{f}_n(X_j)-f_0(X_j)|\}_{j=1}^{10000}$. Based on these errors, we report the estimated $L^2$ and $L^\infty$ errors in Tables~\ref{tab:huber-t2} and \ref{tab:huber-t4} under $t_2$ and $t_4$ noise, respectively. The corresponding side-by-side boxplots for the $L^2$ and $L^\infty$ metrics are presented in Figures~\ref{fig:boxplot of loss, L2} and~\ref{fig:boxplot of loss, Linf}, respectively. As shown in Tables~\ref{tab:huber-t2}--\ref{tab:huber-t4} and Figures~\ref{fig:boxplot of loss, L2}--\ref{fig:boxplot of loss, Linf}, the SiLU FNN and ResNet estimators consistently outperform the ReLU FNN estimator across all settings. Moreover, SiLU ResNet estimators exhibit smaller estimation error compared to SiLU FNN estimators, which is consistent with the optimization advantages of residual architectures. To further assess the statistical estimation performance, we additionally report the estimated convergence exponents. Specifically, under the scaling law $n^{-b}$ for the estimation error, we estimate $b$ via log--log regression and report the results in Tables \ref{tab:huber-t2} and \ref{tab:huber-t4}, together with bootstrap standard deviations based on 1000 resamples. Overall, while all estimators yield comparable exponents for the $L^2$ norm, ReLU FNNs have much smaller uniform convergence exponents. This empirical pattern is consistent with our theory: ReLU FNNs are not well suited to tasks requiring uniform convergence.

\subsection{Real Data Analysis: Temperature Sensitivity of Ozone}\label{sec:realdata_ozone}

Ozone is widely recognized as an air pollutant that affects human health~\cite{schnell2017co}. The IPCC Sixth Assessment Report assesses that climate-driven changes in meteorology can influence ozone concentration, and that shifts in the geographical distribution of emissions have contributed to strong regional variations in ozone trends~\cite{szopa2023short}. These assessments motivate us to study the relationship between temperature and ozone concentrations over the contiguous United States (CONUS) using quantile regression. From a methodological perspective, uniform convergence is important here because it supports reliable estimation uniformly over the spatial domain, rather than only in an average sense.

Our response is the daily maximum 8-hour average ozone concentration, measured in parts per billion (ppb), obtained from the U.S.\ EPA Air Quality System (AQS). The covariates include latitude, longitude, year, day-of-year, and daily maximum and minimum near-surface temperatures (tmmx, tmmn) in $^\circ$C at AQS monitoring sites. The temperature covariates are obtained from the gridMET dataset~\cite{abatzoglou2013development} and mapped to each monitoring site via nearest-neighbor assignment to the closest gridMET grid cell. We construct the dataset using daily observations from the warm season (May--September) over the years 2015--2024, when photochemical ozone formation is most active. The resulting dataset contains approximately 1.84 million observations from about 1{,}500 monitoring sites.

For this quantile regression task, we consider a SiLU ResNet and a ReLU FNN, both with depth 6 and width 128, and train them in PyTorch using the Adam optimizer for 200 epochs. As in Section~\ref{sec: Simulation Experiments}, we take $\Lcal_1$ and $\Lcal_3$ for the SiLU RNs in Definition~\ref{def: Residual block} to be identity maps so that the two architectures have comparable parameter sizes. We fit conditional quantiles at levels $\tau\in\{0.10,0.30,0.50,0.87\}$, following the practice in ozone studies~\cite{schnell2016effect}, since these quantiles admit meaningful practical interpretations. Specifically, the $\tau=0.10$ quantile reflects relatively clean-condition days, the $\tau=0.30$ quantile serves as a robust proxy for the photochemical baseline, the $\tau=0.50$ quantile summarizes the typical ozone level, and the $\tau=0.87$ quantile provides an upper-tail summary of broadly elevated ozone days during the warm season.

\begin{figure}[H]
    \centering
    \includegraphics[width=\textwidth]{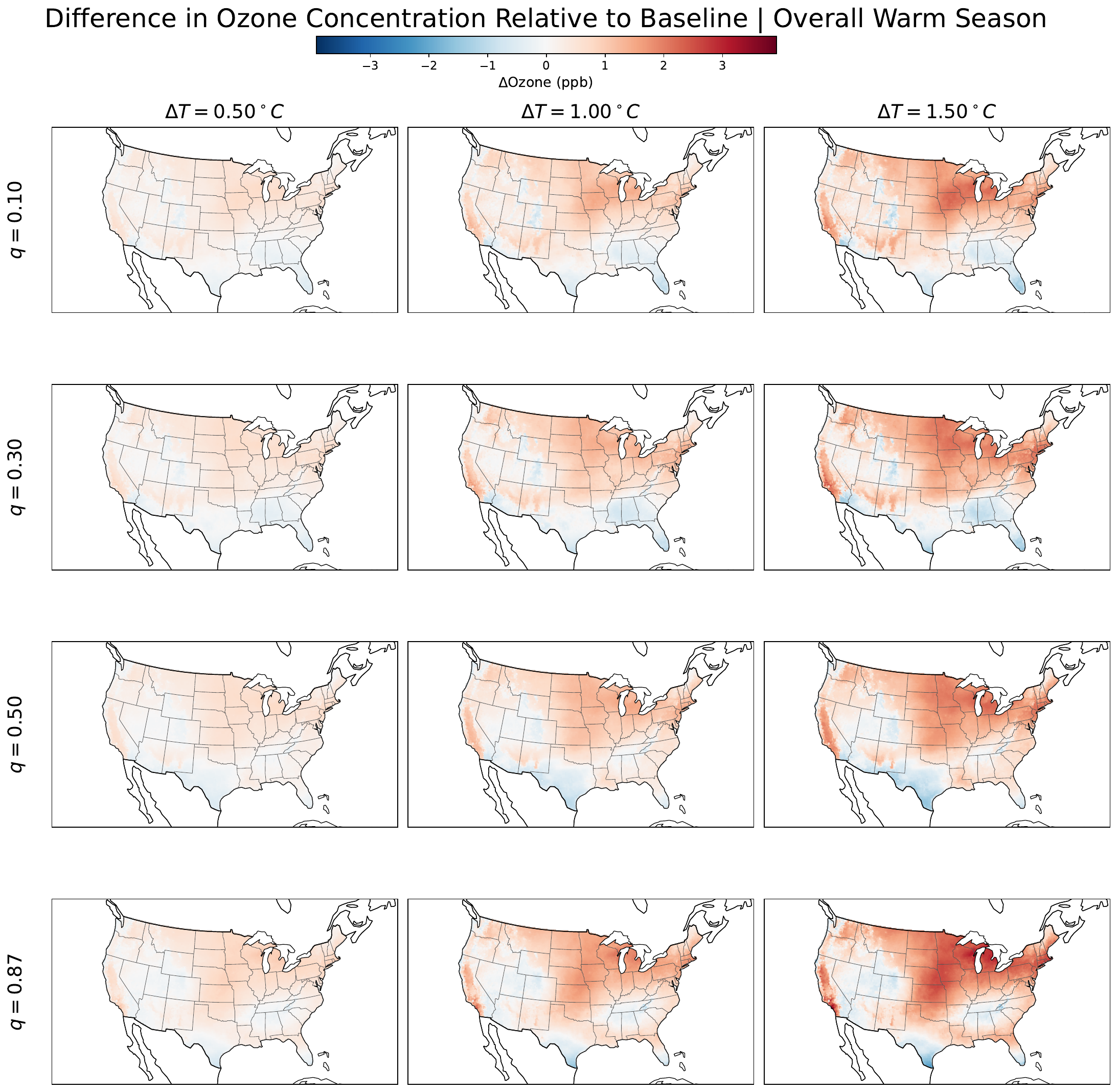}
    \caption{Overall counterfactual difference by SiLU ResNets in ozone concentration.}
    \label{fig: Overall different in ozone concentration relative to baseline}
\end{figure}

We then perform a counterfactual perturbation analysis using the trained models. Specifically, using temperatures from the warm season of 2025, we construct perturbed covariates by adding a constant warming increment $\Delta\in\{0.5,1.0,1.5\}$ to both tmmx and tmmn, while holding the remaining covariates fixed. Using the original and perturbed covariates, we compute predicted daily ozone concentrations and report their differences over the CONUS at 20km resolution. Averaging the predictions over the warm season, Figure~\ref{fig: Overall different in ozone concentration relative to baseline} presents the counterfactual differences in ozone concentrations across quantiles predicted by the SiLU ResNets over the CONUS.

Figure~\ref{fig: Overall different in ozone concentration relative to baseline} reveals a spatially heterogeneous temperature sensitivity of ozone across the CONUS. Across all four quantile levels, warming induces broadly positive changes over the northern tier of the United States and California. These patterns are consistent with prior studies based on classical models~\cite{schnell2016effect, schwarz2021spatial}. In contrast, ozone decreases over parts of the South and the Gulf Coast, consistent with prior evidence that, over tropical and subtropical oceans, increases in temperature and water vapor can reduce photochemical ozone production~\cite{johnson1999relative}. Taken together, these counterfactual patterns suggest that the SiLU ResNet estimators capture climate-driven and geographically heterogeneous temperature--ozone relationships.

To further compare the counterfactual prediction performance of the ReLU FNN and SiLU ResNet estimators, we complement the visual inspection with two quantitative diagnostics. The is \emph{Spatial Roughness} (SR), measures the local irregularity of a rasterized counterfactual ozone-change surface through first-order finite differences; larger values indicate less spatial smoothness and more abrupt local variation. The second is \emph{Extreme Value Fraction} (EVF), defined as the fraction of grid cells with extremely large values. This diagnostic measures the tendency of a fitted model to generate excessive local extremes. For each model, Table~\ref{tab:spatial_diagnostics_ozone} reports the average diagnostic value over all combinations of quantile level $\tau$ and warming increment $\Delta$ within each month, as well as over the full warm season for the overall summary.

More precisely, let $D_{\tau,\Delta,t}(s)$ denote the daily counterfactual ozone change at grid cell $s$, for quantile level $\tau$, warming increment $\Delta$, and date $t$. For each month $m$, and for the overall warm season, we first average $D_{\tau,\Delta,t}(s)$ over time to obtain a mean counterfactual field $Z(s)$. Let $Z$ be a rasterized counterfactual ozone-change field on a two-dimensional grid, represented by the array $(Z_{i,j})_{i,j}$. Let $\Gcal\subseteq\NN_+\times\NN_+$ denote the set of valid grid-cell indices, and let $\Hcal,\Vcal\subseteq\NN_+\times\NN_+$ denote the sets of valid grid cells whose horizontal and vertical neighbors are also valid, respectively, namely,
\[
    (i,j)\in \Hcal \text{ if } (i,j)\in\Gcal \text{ and } (i,j+1)\in\Gcal;\qquad
    (i,j)\in \Vcal \text{ if } (i,j)\in\Gcal \text{ and } (i+1,j)\in\Gcal.
\]
We define the spatial roughness of $Z$ by
\[
    \mathrm{SR}(Z)= \frac{\sum_{(i,j)\in \Hcal}|Z_{i,j+1}-Z_{i,j}|+\sum_{(i,j)\in \Vcal}|Z_{i+1,j}-Z_{i,j}|}{|\Hcal|+|\Vcal|},
\]
that is, the mean absolute difference over all valid horizontal and vertical neighboring pairs, and
\[
    \mathrm{EVF}(Z;u)= \frac{1}{|\Gcal|}\sum_{s\in\Gcal} \mathbf{1}\{|Z(s)|>u\},
\]
with threshold $u=3.0$ ppb. For each model and each month, we average these diagnostics over all quantile levels $\tau\in\{0.10,0.30,0.50,0.87\}$ and warming increments $\Delta\in\{0.5,1.0,1.5\}$. Standard deviations are computed using a month-stratified paired bootstrap over dates with 1000 resamples.

\begin{table}[h]
\caption{Diagnostics for the counterfactual ozone-change predictions of the ReLU FNNs and SiLU ResNets, with bootstrap standard deviations in parentheses.}
    \label{tab:spatial_diagnostics_ozone}
    \centering
\scalebox{1.0}{
\begin{tabular}{c|cc|cc}
\hline
Metric & \multicolumn{2}{c|}{SR} & \multicolumn{2}{c}{EVF} \\ \hline
Model & ReLU FNN & SiLU RN & ReLU FNN & SiLU RN \\
\hline
May & 0.1174 (0.0156) & \textbf{0.0705 (0.0069)} & 0.0011 (0.0022) & \textbf{0.0007 (0.0008)} \\
June & 0.1564 (0.0181) & \textbf{0.0768 (0.0063)} & 0.0137 (0.0039) & \textbf{0.0020 (0.0029)} \\
July & 0.1589 (0.0165) & \textbf{0.0918 (0.0054)} & 0.0227 (0.0060) & \textbf{0.0097 (0.0041)} \\
August & 0.1448 (0.0177) & \textbf{0.0804 (0.0070)} & \textbf{0.0107 (0.0030)} & 0.0108 (0.0018) \\
September & 0.1409 (0.0194) & \textbf{0.0759 (0.0066)} & 0.0085 (0.0043) & \textbf{0.0047 (0.0022)} \\
Overall & 0.0813 (0.0037) & \textbf{0.0555 (0.0017)} & 0.0012 (0.0008) & \textbf{0.0004 (0.0006)} \\
\hline
\end{tabular}
}
\end{table}

As shown in Table~\ref{tab:spatial_diagnostics_ozone}, the SiLU ResNet yields uniformly smaller SR values than the ReLU FNN, indicating more spatially coherent counterfactual predictions. For EVF, the SiLU ResNet also attains smaller values in most months and in the overall summary. In addition, the SiLU ResNet exhibits smaller bootstrap standard deviations for both diagnostics in all reported cases. Taken together, these results suggest that the SiLU ResNet produces more stable and spatially coherent quantile predictions, while generating fewer potentially spurious extreme local responses in this application.

\section{Conclusion}\label{sec: Conclusion}

In this work, we investigate the slow uniform convergence of ReLU FNN estimators and develop a statistical theory for smooth DNNs, an alternative class that aligns with modern deep learning practice while enjoying uniform convergence guarantees in many statistical learning tasks. We first show that, despite minimax-optimal $L^2(P)$ convergence, least-squares ReLU FNN estimators can suffer from the \textit{curse of dimensionality} in uniform convergence, owing to the low smoothness of the ReLU activation. Motivated by the prevalence of residual architectures and $C^\infty$ activations in state-of-the-art deep learning models, we introduce smooth DNNs and establish their core statistical properties, including pseudo-dimension bounds, approximation error bounds for Sobolev functions and hierarchical composition models, and factorial-type Hölder norm bounds for approximators. Building on these foundations, we derive non-asymptotic $L^\infty$ convergence rates for smooth DNN estimators in several representative tasks: Huber, least-squares, quantile, and logistic regression, showing that smooth DNNs can provide reliable uniform convergence guarantees while adapting to low-dimensional hierarchical structure. Together, these results position smooth DNNs as a theoretically grounded alternative to ReLU FNN estimators in settings where uniform convergence is essential.

\printbibliography

@article{sun2020adaptive,
  title={Adaptive huber regression},
  author={Sun, Qiang and Zhou, Wen-Xin and Fan, Jianqing},
  journal={Journal of the American Statistical Association},
  volume={115},
  number={529},
  pages={254--265},
  year={2020},
  publisher={Taylor \& Francis}
}

@article{gagliardo1959ulteriori,
  title={Ulteriori propriet{\`a} di alcune classi di funzioni in pi{\`u} variabili},
  author={Gagliardo, Emilio},
  journal={Ricerche Mat.},
  volume={8},
  pages={24--51},
  year={1959}
}

@article{nirenberg1966extended,
  title={An extended interpolation inequality},
  author={Nirenberg, Louis},
  journal={Annali della Scuola Normale Superiore di Pisa-Scienze Fisiche e Matematiche},
  volume={20},
  number={4},
  pages={733--737},
  year={1966}
}

@article{kuchibhotla2022least,
  title={On least squares estimation under heteroscedastic and heavy-tailed errors},
  author={Kuchibhotla, Arun K and Patra, Rohit K},
  journal={The Annals of Statistics},
  volume={50},
  number={1},
  pages={277--302},
  year={2022},
  publisher={Institute of Mathematical Statistics}
}

@article{han2019convergence,
author = {Qiyang Han and Jon A. Wellner},
title = {{Convergence rates of least squares regression estimators with heavy-tailed errors}},
volume = {47},
journal = {The Annals of Statistics},
number = {4},
publisher = {Institute of Mathematical Statistics},
pages = {2286 -- 2319},
year = {2019}
}

@article{brezis2018gagliardo,
  title={Gagliardo--Nirenberg inequalities and non-inequalities: the full story},
  author={Brezis, Haïm and Mironescu, Petru},
  journal={Annales de l'Institut Henri Poincar{\'e} C, Analyse non lin{\'e}aire},
  volume={35},
  number={5},
  pages={1355--1376},
  year={2018},
  publisher={Elsevier}
}

@article{fan2024noise,
  title={How do noise tails impact on deep ReLU networks?},
  author={Fan, Jianqing and Gu, Yihong and Zhou, Wen-Xin},
  journal={The Annals of Statistics},
  volume={52},
  number={4},
  pages={1845--1871},
  year={2024},
  publisher={Institute of Mathematical Statistics}
}

@article{schmidt2020nonparametric,
author = {Johannes Schmidt-Hieber},
title = {Nonparametric regression using deep neural networks with ReLU activation function},
volume = {48},
journal = {The Annals of Statistics},
number = {4},
publisher = {Institute of Mathematical Statistics},
pages = {1875 -- 1897},
year = {2020}
}

@article{yarotsky2017error,
  title={Error bounds for approximations with deep ReLU networks},
  author={Yarotsky, Dmitry},
  journal={Neural Networks},
  volume={94},
  pages={103--114},
  year={2017},
  publisher={Elsevier}
}

@article{lu2021deep,
  title={Deep network approximation for smooth functions},
  author={Lu, Jianfeng and Shen, Zuowei and Yang, Haizhao and Zhang, Shijun},
  journal={SIAM Journal on Mathematical Analysis},
  volume={53},
  number={5},
  pages={5465--5506},
  year={2021},
  publisher={SIAM}
}

@article{kohler2021rate,
  title={On the rate of convergence of fully connected deep neural network regression estimates},
  author={Kohler, Michael and Langer, Sophie},
  journal={The Annals of Statistics},
  volume={49},
  number={4},
  pages={2231--2249},
  year={2021},
  publisher={Institute of Mathematical Statistics}
}

@article{bartlett2019nearly,
  title={Nearly-tight VC-dimension and pseudodimension bounds for piecewise linear neural networks},
  author={Bartlett, Peter L and Harvey, Nick and Liaw, Christopher and Mehrabian, Abbas},
  journal={Journal of Machine Learning Research},
  volume={20},
  number={63},
  pages={1--17},
  year={2019}
}

@article{bauer2019deep,
  title={On deep learning as a remedy for the curse of dimensionality in nonparametric regression},
  author={Bauer, Benedikt and Kohler, Michael},
  year={2019},
  volume = {47},
  journal = {The Annals of Statistics},
  number = {4},
  publisher = {Institute of Mathematical Statistics},
  pages = {2261 -- 2285}
}

@book{anthony2009neural,
  title={Neural Network Learning: Theoretical Foundations},
  author={Anthony, Martin and Bartlett, Peter L},
  year={2009},
  publisher={Cambridge University Press}
}

@book{koenker2005quantile,
  title={Quantile Regression},
  author={Koenker, Roger},
  volume={38},
  year={2005},
  publisher={Cambridge University Press}
}

@article{zhong2022deep,
  title={Deep learning for the partially linear Cox model},
  author={Zhong, Qixian and Mueller, Jonas and Wang, Jane-Ling},
  journal={The Annals of Statistics},
  volume={50},
  number={3},
  pages={1348--1375},
  year={2022},
  publisher={Institute of Mathematical Statistics}
}

@article{fan2024factor,
  title={Factor augmented sparse throughput deep relu neural networks for high dimensional regression},
  author={Fan, Jianqing and Gu, Yihong},
  journal={Journal of the American Statistical Association},
  volume={119},
  number={548},
  pages={2680--2694},
  year={2024},
  publisher={Taylor \& Francis}
}

@article{bhattacharya2024deep,
  title={Deep neural networks for nonparametric interaction models with diverging dimension},
  author={Bhattacharya, Sohom and Fan, Jianqing and Mukherjee, Debarghya},
  journal={The Annals of Statistics},
  volume={52},
  number={6},
  pages={2738--2766},
  year={2024},
  publisher={Institute of Mathematical Statistics}
}

@article{yao2005functional,
  title = {Functional linear regression analysis for longitudinal data},
  author = {Fang Yao and Hans‐Georg Müller and Jane‐Ling Wang},
  journal = {The Annals of Statistics},
  volume = {33},
  number = {6},
  pages = {2873--2903},
  year = {2005}
}

@article{guo2021estimation,
  title={Estimation of optimal individualized treatment rules using a covariate-specific treatment effect curve with high-dimensional covariates},
  author={Guo, Wenchuan and Zhou, Xiao-Hua and Ma, Shujie},
  journal={Journal of the American Statistical Association},
  volume={116},
  number={533},
  pages={309--321},
  year={2021},
  publisher={Taylor \& Francis}
}

@article{schmidt2024local,
  title={Local convergence rates of the nonparametric least squares estimator with applications to transfer learning},
  author={Schmidt-Hieber, Johannes and Zamolodtchikov, Petr},
  journal={Bernoulli},
  volume={30},
  number={3},
  pages={1845--1877},
  year={2024},
  publisher={Bernoulli Society for Mathematical Statistics and Probability}
}

@article{karpinski1997polynomial,
  title={Polynomial bounds for VC dimension of sigmoidal and general Pfaffian neural networks},
  author={Karpinski, Marek and Macintyre, Angus},
  journal={Journal of Computer and System Sciences},
  volume={54},
  number={1},
  pages={169--176},
  year={1997},
  publisher={Elsevier}
}

@inproceedings{he2015resnet,
  title={Deep residual learning for image recognition},
  author={He, Kaiming and Zhang, Xiangyu and Ren, Shaoqing and Sun, Jian},
  booktitle={Proceedings of the IEEE Conference on Computer Vision and Pattern Recognition},
  pages={770--778},
  year={2016}
}

@article{ramachandran2017swish,
  title={Searching for activation functions},
  author={Ramachandran, Prajit and Zoph, Barret and Le, Quoc V},
  journal={arXiv preprint arXiv:1710.05941},
  year={2017}
}

@article{elfwing2018silu,
  title={Sigmoid-weighted linear units for neural network function approximation in reinforcement learning},
  author={Elfwing, Stefan and Uchibe, Eiji and Doya, Kenji},
  journal={Neurocomputing},
  volume={275},
  pages={2166--2173},
  year={2018},
  publisher={Elsevier}
}

@article{hendrycks2016gelu,
  title={Gaussian error linear units (GELUs)},
  author={Hendrycks, Dan and Gimpel, Kevin},
  journal={arXiv preprint arXiv:1606.08415},
  year={2016}
}

@inproceedings{misra2019mish,
  title={Mish: A self regularized non-monotonic neural activation function},
  author={Misra, Diganta},
  booktitle={Proceedings of the British Machine Vision Conference 2020},
  year={2020}
}

@article{touvron2023llama,
  title        = {LLaMA: Open and Efficient Foundation Language Models},
  author       = {Touvron, Hugo and Lavril, Thibaut and Izacard, Gautier and Martinet, Xavier and Lachaux, Marie-Anne and Lacroix, Timoth{\'e}e and Rozi{\`e}re, Baptiste and Goyal, Naman and Hambro, Eric and Azhar, Faisal and Rodriguez, Aurelien and Joulin, Armand and Grave, Edouard and Lample, Guillaume},
  journal      = {arXiv preprint arXiv:2302.13971},
  year         = {2023}
}

@inproceedings{dosovitskiy2021vit,
  title        = {An Image is Worth 16x16 Words: Transformers for Image Recognition at Scale},
  author       = {Dosovitskiy, Alexey and Beyer, Lucas and Kolesnikov, Alexander and Weissenborn, Dirk and Zhai, Xiaohua and Unterthiner, Thomas and Dehghani, Mostafa and Minderer, Matthias and Heigold, Georg and Gelly, Sylvain and Uszkoreit, Jakob and Houlsby, Neil},
  booktitle    = {International Conference on Learning Representations (ICLR)},
  year         = {2021}
}

@article{ding2025new,
  title={New Empirical Process Tools and Their Applications to Robust Deep ReLU Networks and Phase Transitions for Nonparametric Regression},
  author={Ding, Yizhe and Li, Runze and Xue, Lingzhou},
  journal={arXiv preprint arXiv:2511.15841},
  year={2025}
}

@inproceedings{he2016identity,
  title     = {Identity Mappings in Deep Residual Networks},
  author    = {He, Kaiming and Zhang, Xiangyu and Ren, Shaoqing and Sun, Jian},
  booktitle = {European Conference on Computer Vision (ECCV)},
  pages     = {630--645},
  publisher = {Springer},
  year      = {2016}
}

@inproceedings{goldberg1993bounding,
  title={Bounding the Vapnik-Chervonenkis dimension of concept classes parameterized by real numbers},
  author={Goldberg, Paul and Jerrum, Mark},
  booktitle={Proceedings of the sixth Annual Conference on Computational Learning Theory},
  pages={361--369},
  year={1993}
}

@article{shen2023differentiable,
  title={Differentiable neural networks with RePU activation: With applications to score estimation and isotonic regression},
  author={Shen, Guohao and Jiao, Yuling and Lin, Yuanyuan and Huang, Jian},
  journal={arXiv preprint arXiv:2305.00608},
  year={2023}
}

@article{belomestny2023simultaneous,
  title={Simultaneous approximation of a smooth function and its derivatives by deep neural networks with piecewise-polynomial activations},
  author={Belomestny, Denis and Naumov, Alexey and Puchkin, Nikita and Samsonov, Sergey},
  journal={Neural Networks},
  volume={161},
  pages={242--253},
  year={2023},
  publisher={Elsevier}
}

@article{de2021approximation,
  title={On the approximation of functions by tanh neural networks},
  author={De Ryck, Tim and Lanthaler, Samuel and Mishra, Siddhartha},
  journal={Neural Networks},
  volume={143},
  pages={732--750},
  year={2021},
  publisher={Elsevier}
}

@article{zhang2024deep,
  title={Deep network approximation: Beyond relu to diverse activation functions},
  author={Zhang, Shijun and Lu, Jianfeng and Zhao, Hongkai},
  journal={Journal of Machine Learning Research},
  volume={25},
  number={35},
  pages={1--39},
  year={2024}
}

@article{yang2025deep,
  title={Deep Neural Networks with General Activations: Super-Convergence in Sobolev Norms},
  author={Yang, Yahong and He, Juncai},
  journal={arXiv preprint arXiv:2508.05141},
  year={2025}
}

@article{xie2025mhc,
  title={mhc: Manifold-constrained hyper-connections},
  author={Xie, Zhenda and Wei, Yixuan and Cao, Huanqi and Zhao, Chenggang and Deng, Chengqi and Li, Jiashi and Dai, Damai and Gao, Huazuo and Chang, Jiang and Zhao, Liang and others},
  journal={arXiv preprint arXiv:2512.24880},
  year={2025}
}

@article{yu2025deep,
  title={Deep neural expected shortfall regression with tail-robustness},
  author={Yu, Myeonghun and Tan, Kean Ming and Wang, Huixia Judy and Zhou, Wen-Xin},
  journal={arXiv preprint arXiv:2511.08772},
  year={2025}
}

@article{schnell2016effect,
  title   = {Effect of climate change on surface ozone over North America, Europe, and East Asia},
  author  = {Schnell, Jordan L. and Prather, Michael J. and Josse, Beatrice and Naik, Vaishali and Horowitz, Larry W. and Zeng, Guohui and Shindell, Drew T. and Faluvegi, Greg and Stegmann, Petra and Folberth, Gerd A. and others},
  journal = {Geophysical Research Letters},
  year    = {2016},
  volume  = {43},
  number  = {7},
  pages   = {3509--3518}
}

@article{szopa2023short,
  title={Short-Lived Climate Forcers (Chapter 6)},
  author={Szopa, Sophie and Naik, Vaishali and Adhikary, Bhupesh and Artaxo, Paulo and Berntsen, Terje and Collins, William Drew and Fuzzi, Sandro and Gallardo, Laura and Kiendler-Scharr, Astrid and Klimont, Zbigniew and others},
  journal={Intergovernmental Panel on Climate Change (IPCC) 2021: Climate Change 2021: The Physical Science Basis.},
  pages={817--922},
  year={2023},
  publisher={Cambridge University Press}
}

@article{abatzoglou2013development,
  title={Development of gridded surface meteorological data for ecological applications and modelling},
  author={Abatzoglou, John T},
  journal={International Journal of Climatology},
  volume={33},
  number={1},
  pages={121--131},
  year={2013},
  publisher={John Wiley \& Sons, Ltd. Chichester, UK}
}

@article{johnson1999relative,
  title={Relative roles of climate and emissions changes on future tropospheric oxidant concentrations},
  author={Johnson, CE and Collins, WJ and Stevenson, DS and Derwent, RG},
  journal={Journal of Geophysical Research: Atmospheres},
  volume={104},
  number={D15},
  pages={18631--18645},
  year={1999},
  publisher={Wiley Online Library}
}

@article{schnell2017co,
  title={Co-occurrence of extremes in surface ozone, particulate matter, and temperature over eastern North America},
  author={Schnell, Jordan L and Prather, Michael J},
  journal={Proceedings of the National Academy of Sciences},
  volume={114},
  number={11},
  pages={2854--2859},
  year={2017},
  publisher={National Academy of Sciences}
}

@article{schwarz2021spatial,
  title={Spatial variation in the joint effect of extreme heat events and ozone on respiratory hospitalizations in California},
  author={Schwarz, Lara and Hansen, Kristen and Alari, Anna and Ilango, Sindana D and Bernal, Nelson and Basu, Rupa and Gershunov, Alexander and Benmarhnia, Tarik},
  journal={Proceedings of the National Academy of Sciences},
  volume={118},
  number={22},
  pages={e2023078118},
  year={2021},
  publisher={National Academy of Sciences}
}

@article{imaizumi2023sup,
  title={Sup-norm convergence of deep neural network estimator for nonparametric regression by adversarial training},
  author={Imaizumi, Masaaki},
  journal={arXiv preprint arXiv:2307.04042},
  year={2023}
}

@article{manole2024background,
  title={Background modeling for double Higgs boson production: Density ratios and optimal transport},
  author={Manole, Tudor and Bryant, Patrick and Alison, John and Kuusela, Mikael and Wasserman, Larry},
  journal={The Annals of Applied Statistics},
  volume={18},
  number={4},
  pages={2950--2978},
  year={2024},
  publisher={Institute of Mathematical Statistics}
}

@inproceedings{oono2019approximation,
  title={Approximation and non-parametric estimation of ResNet-type convolutional neural networks},
  author={Oono, Kenta and Suzuki, Taiji},
  booktitle={International Conference on Machine Learning},
  pages={4922--4931},
  year={2019},
  organization={PMLR}
}

@inproceedings{liu2021besov,
  title={Besov function approximation and binary classification on low-dimensional manifolds using convolutional residual networks},
  author={Liu, Hao and Chen, Minshuo and Zhao, Tuo and Liao, Wenjing},
  booktitle={International Conference on Machine Learning},
  pages={6770--6780},
  year={2021},
  organization={PMLR}
}

@inproceedings{liu2022benefits,
  title={Benefits of overparameterized convolutional residual networks: Function approximation under smoothness constraint},
  author={Liu, Hao and Chen, Minshuo and Er, Siawpeng and Liao, Wenjing and Zhang, Tong and Zhao, Tuo},
  booktitle={International Conference on Machine Learning},
  pages={13669--13703},
  year={2022},
  organization={PMLR}
}

@inproceedings{jiang2017uniform,
  title={Uniform convergence rates for kernel density estimation},
  author={Jiang, Heinrich},
  booktitle={International Conference on Machine Learning},
  pages={1694--1703},
  year={2017},
  organization={PMLR}
}

\end{document}